\documentclass[sigconf, screen]{acmart}
\renewcommand\footnotetextcopyrightpermission[1]{} 
\AtBeginDocument{%
  }

\setcopyright{acmlicensed}
\copyrightyear{2018}
\acmYear{2018}
\acmDOI{XXXXXXX.XXXXXXX}
\acmConference[Conference acronym 'XX]{Make sure to enter the correct
  conference title from your rights confirmation email}{June 03--05,
  2018}{Woodstock, NY}
\acmISBN{978-1-4503-XXXX-X/2018/06}

\usepackage{makecell}
\usepackage{multirow}
\usepackage{booktabs}
\usepackage{graphicx} 

\begin{document}

\title{Making Image Editing Easier via Adaptive Task Reformulation
with Agentic Executions}

\author{Bo Zhao}
\authornote{Both authors contributed equally to this research.}
\email{zhaobo20000116@gmail.com}
\affiliation{%
  \institution{Nanjing University}
  \country{China}
}

\author{Kairui Guo}
\authornotemark[1]
\email{carryguo02@gmail.com}
\affiliation{%
  \institution{Beijing Institute of Technology}
  \country{China}
}

\author{Runnan Du}
\affiliation{%
  \institution{College of Artificial Intelligence, China University of Petroleum (Beijing)}
  \country{China}
}

\author{Haiyang Sun}
\affiliation{%
  \institution{Beijing University of Posts and Telecommunications}
  \country{China}
}

\author{Pengshan Wang}
\affiliation{%
  \institution{Beijing Institute of Technology}
  \country{China}
}

\author{Huan Yang}
\email{hyang@fastmail.com}
\authornote{Corresponding authors}
\affiliation{%
  \institution{Kuaishou Technology}
  \country{China}
}

\author{Kun Gai}
\affiliation{%
  \institution{Kuaishou Technology}
  \country{China}
}

\author{Yixin Cao}
\affiliation{%
  \institution{Fudan University}
  \country{China}
}

\author{Wei Ji}
\email{weiji@nju.edu.cn}
\authornotemark[2]
\affiliation{%
  \institution{Nanjing University}
  \country{China}
}

\renewcommand{\shortauthors}{Trovato et al.}

\begin{abstract}
Instruction-guided image editing has advanced substantially with recent generative models, yet it still fails to produce reliable results across many seemingly simple cases. We observe that a large portion of these failures stem not from insufficient model capacity, but from poorly formulated editing tasks, such as those involving small targets, implicit spatial relations, or under-specified instructions.
In this work, we frame image editing failures as a task formulation problem and propose an adaptive task reformulation framework that improves editing performance without modifying the underlying model. Our key idea is to transform the original image–instruction pair into a sequence of operations that are dynamically determined and executed by a MLLM agent through analysis, routing, reformulation, and feedback-driven refinement.
Experiments on multiple benchmarks, including ImgEdit, PICA, and RePlan, across diverse editing backbones such as Qwen-Image-Edit and Nano Banana, show consistent improvements, with especially large gains on challenging cases. These results suggest that task reformulation is a critical but underexplored factor, and that substantial gains can be achieved by better matching editing tasks to the effective operating regime of existing models. The code is available at \href{https://github.com/KlingAIResearch/ATR}{https://github.com/KlingAIResearch/ATR}.
\end{abstract}

\begin{teaserfigure}
  \includegraphics[width=\textwidth]{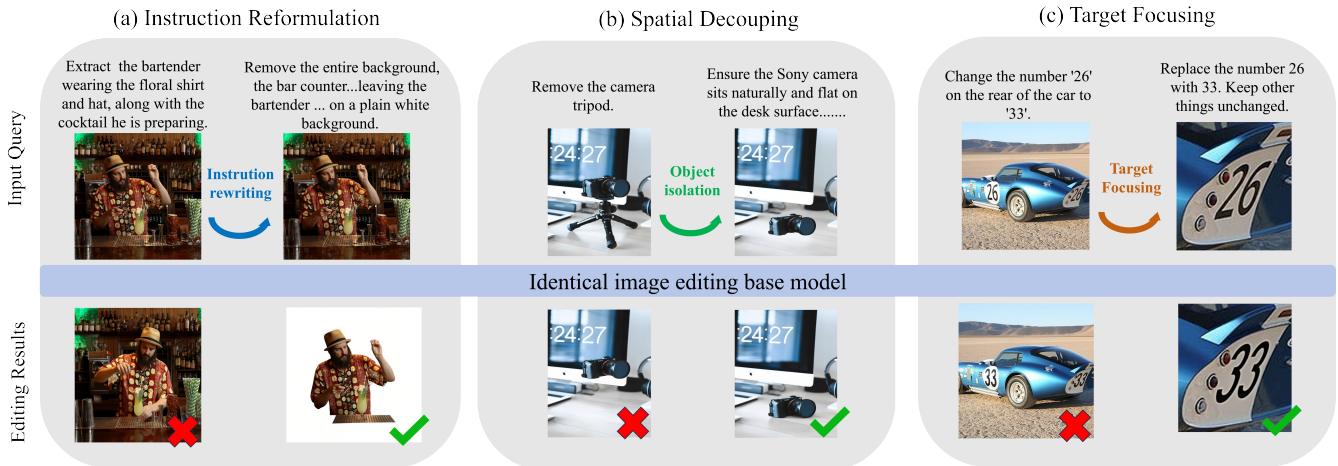}
  \caption{Overcoming editing failures via task reformulation. We show that success depends heavily on task presentation. By keeping the backbone fixed and applying our three strategies, we successfully correct direct editing failures, proving the critical role of input conditioning.}
  \Description{Enjoying the baseball game from the third-base
  seats. Ichiro Suzuki preparing to bat.}
  \label{fig:teaser}
\end{teaserfigure}


\maketitle

\section{Introduction}
Instruction-guided image editing has emerged as a core capability of modern multimodal generative models, enabling users to modify images through natural language. With recent advances in large-scale vision-language modeling and image generation, substantial progress has been made in producing visually plausible and semantically aligned edits from an input image and a textual instruction\cite{brooks2023instructpix2pix, zhang2023magicbrush}. This capability has broad relevance to digital content creation, design assistance and layout generation~\cite{wang2025sega}, personalized media generation, and interactive visual editing systems\cite{pan2023drag, ruiz2023dreambooth}.

Despite this progress, current image editing models still fail frequently on tasks that are, in principle, well within their underlying capabilities. For example, changing the number on an object is often straightforward when the target is large and clearly specified, yet the same operation becomes unreliable when the object is small, embedded in a cluttered scene, or only weakly grounded by the instruction, as shown in Figure~\ref{fig:teaser}. Similar issues arise when edits depend on implicit spatial relations, involve strong local entanglement with surrounding content\cite{chen2025regione}, or require preserving hidden structural dependencies and local textures\cite{tumanyan2023plug,zhang2023adding,zhao2026texeditor} during manipulation. These failures are not isolated corner cases: they occur consistently across standard benchmarks and across different editing backbones, revealing a systematic gap between the reliability of current editing systems and real-world user expectations\cite{basu2023editval,huang2023t2i}.

We argue that this gap is not best understood as a pure model-capacity problem. Instead, many image editing failures arise because the original image--instruction pair is poorly conditioned for direct execution. Existing editing models often operate reliably only when certain implicit conditions are satisfied: the target should be sufficiently prominent, the spatial grounding should be clear, and the intended operation should be adequately specified, as shown in Figure~\ref{fig:teaser}. When these conditions are not met, directly applying the editing model to the raw query often leads to suboptimal results, even though the desired edit itself is not inherently difficult. This suggests that editing success depends not only on what the model can do, but also on how the task is formulated for the model.

Motivated by this observation, we revisit instruction-guided image editing from the perspective of task reformulation. Rather than modifying the backbone model\cite{zhang2023sine, kawar2023imagic, fu2023guiding}, we aim to transform the original editing query into a form that is better aligned with the model’s effective operating conditions. From this perspective, common failures—such as ambiguous targeting, poor local focus, or structural violations—stem from a task-model mismatch rather than inherent model weaknesses. Our core idea is therefore to reformulate a single difficult editing problem into a sequence of more structured and better-conditioned operations, so that challenging edits can be resolved more robustly without changing the underlying model.

Based on this perspective, we propose an adaptive task reformulation framework implemented by a lightweight agent with multi-step, feedback-driven decision making at inference time. Given an input image and editing instruction, the agent first performs query profiling to characterize the task in terms of its edit target, key constraints, and scope of modification, while extracting target-aware visual context and relevant spatial properties such as relative scale, position, and surrounding dependencies. It then performs reformulation routing to assign the query to a suitable execution strategy, such as direct editing, structural decoupling, or localized crop editing. Within the selected route, the agent carries out route-conditioned execution by iteratively invoking reformulation operations and editing tools to manipulate the task representation, for example by isolating a target region, decoupling structurally dependent content, rewriting ambiguous intent, or composing localized edits back into the original image. Intermediate outputs are further used as feedback to guide subsequent planning and action selection, allowing the system to progressively refine the editing process rather than relying on a single-shot prediction.

We evaluate our approach on multiple image editing benchmarks, including ImgEdit\cite{ye2025imgedit}, PICA\cite{pu2025picabench}, and RePlan\cite{qu2025replan}, across diverse editing backbones such as Qwen-Image-Edit and Nano Banana. Our framework consistently improves performance across benchmarks and models, with especially pronounced gains on challenging subsets associated with common failure modes. These results show that task reformulation is a critical yet largely underexplored dimension of image editing, complementary to advances in model architecture and training. More broadly, they suggest that a substantial portion of current editing failures can be mitigated not by building a new model, but by better organizing the task presented to existing ones. 
Our contributions are three-fold:
\begin{itemize}
    \item We identify task formulation mismatch as a key and underexplored source of failure in instruction-guided image editing, and argue that many unreliable edits stem from poorly conditioned queries rather than insufficient model capability.
    
    \item We propose an adaptive task reformulation framework that profiles each query, routes it to a suitable reformulation strategy, and executes it through route-conditioned agentic operations with intermediate states and feedback.
    
    \item We demonstrate consistent improvements across multiple benchmarks and backbone models, showing that task reformulation provides a practical and model-agnostic path toward more reliable image editing.
\end{itemize}

\section{Related Work}  
\subsection{Instruction-guided Image Editing}

Instruction-guided image editing has progressed rapidly with diffusion models and multimodal generative systems, building upon foundational advances in image completion~\cite{zhao2025learning} and manipulation\cite{couairon2022diffedit, zhou2025fireedit}. Early work such as InstructPix2Pix~\cite{brooks2023instructpix2pix} established the paradigm of directly mapping an image--instruction pair to an edited result, while more recent systems such as SD3.5~\cite{esser2024scaling}, Emu Edit \cite{sheynin2024emu}, SmartEdit \cite{huang2024smartedit}, and Qwen-Image-Edit~\cite{wu2025qwen} further improve fidelity and instruction following through stronger model capacity and alignment.  

However, challenging cases remain, including small targets, implicit spatial relations, and under-specified instructions~\cite{basu2023editval}. Most existing approaches address these issues by improving the editor itself, for example through larger models, better data, or stronger multimodal alignment~\cite{podell2023sdxl, betker2023improving, geng2024instructdiffusion}. In contrast, we emphasize that many failures stem from how the task is formulated, rather than only from limitations of the backbone model.

\subsection{Prompt Engineering and Input Reformulation}

Another related direction improves generation through prompt engineering and input reformulation~\cite{hao2023optimizing, lian2023llm, liu2022design}. Industrial systems and prompt optimization methods show that instruction clarity and completeness can strongly affect model behavior without changing model parameters. EditThinker further extends this idea by introducing intermediate reasoning before editing~\cite{li2025editthinker}.  

Nevertheless, these methods mainly focus on the textual side of the task, such as rewriting or reasoning over instructions, while leaving the visual execution process largely unchanged~\cite{cao2023beautifulprompt, yang2024mastering}. Our method instead jointly reformulates the task and adapts the execution strategy, enabling better alignment between task specification and model behavior.

\subsection{Agent-based and Planning-based Image Editing}

Recent methods introduce agents and planning mechanisms for multi-step image editing.Inspired by early successes of LLMs serving as task routers and tool executors \cite{shen2023hugginggpt}, GenArtist~\cite{wang2024genartist}, RePlan~\cite{qu2025replan}, and EditThinker~\cite{li2025editthinker} improve editing through reasoning, decomposition, and tool use. However, they are primarily execution-oriented and generally assume the original task specification is already appropriate.  

By contrast, we view image editing as a sequential decision-making problem that jointly involves task reformulation and execution. Our framework first transforms the task according to input characteristics, and then selects suitable execution strategies through routing.

\section{Pilot Study: from failure cases to task reformulation }

As illustrated in Figure~\ref{fig:teaser}, even when two editing tasks are semantically close, directly solving them in the original formulation can lead to substantially different outcomes. In contrast, by reformulating the task or transforming the input into a more favorable form, the same editing model can achieve a much higher success rate. This observation suggests that a nontrivial portion of editing failures may arise not from insufficient model capacity, but from a mismatch between the input formulation and the model's effective operating regime. However, whether this phenomenon is general and practically useful on real benchmark cases remains unclear.

To examine this, we conduct a pilot study on challenging real-world editing examples collected from PICA and ImgEdit Bench. Specifically, we select the bottom 40\% cases where Qwen-Image-Edit performs poorly under direct editing, and use them as a testbed for analyzing whether task reformulation can systematically recover performance. We first employ an MLLM to inspect the failed direct-editing results and provide failure attributions. These attributions are then consolidated into the seven categories listed on the left side of Table~\ref{tab:bad_case_analysis}. The result trend in ImgEdit Bench is similar and  more details of the pilot study can be found in Appendix.

For each failure category, we compare four editing strategies: direct editing, instruction rewriting, target-region cropping, and spatial disentanglement of the target object. The results reveal three key findings:

\begin{enumerate}
    \item For each failure category, there exists at least one reformulation strategy that consistently outperforms direct editing.
    
    \item Different failure subsets benefit from different forms of reformulation, indicating that there is no single universally optimal transformation.
    
    \item The majority of failure cases can be effectively covered by our taxonomy and the corresponding reformulation strategies.
\end{enumerate}

These findings provide empirical support for our central hypothesis: many hard editing cases can be converted into easier ones by selecting an appropriate reformulation before execution. This, in turn, suggests a clear path toward more reliable image editing: if we can build an automatic system that adaptively decides when reformulation is needed, selects the appropriate reformulation strategy for each input case, and executes the transformed editing process robustly, then stable performance gains can be achieved beyond naive direct editing.

\begin{table}[t]
\centering
\caption{Failure-case analysis across different reasoning strategies.}
\label{tab:bad_case_analysis}
\renewcommand{\arraystretch}{1.12}
\resizebox{\columnwidth}{!}{%
\begin{tabular}{lccccc}
\toprule
\multirow{2}{*}{Failure category} & \multirow{2}{*}{\makecell{Count \\ (\%)}} & \multicolumn{4}{c}{Reasoning strategy} \\
\cmidrule(lr){3-6}
& & Direct & Rewrite & Spatial & Local \\
\midrule
Target ambiguity              & 9  (4.6\%)  & 21.48 & 43.89 & 48.96 & \textbf{53.12} \\
Local entanglement            & 12 (6.2\%)  & 36.67 & 34.69 & 43.76 & \textbf{52.75} \\
Structural dependency         & 31 (16.0\%) & 31.85 & 45.73 & \textbf{53.22} & 42.63 \\
Hidden-content reconst. & 32 (16.5\%) & 36.51 & 37.83 & 35.00 & \textbf{44.46} \\
Scene-wide consistency        & 78 (40.2\%) & 40.87 & \textbf{53.47} & 53.27 & 48.15 \\
Transformation complexity     & 31 (16.0\%) & 36.86 & \textbf{50.84} & 45.61 & 35.59 \\
Others                        & 1  (0.5\%)  & \textbf{33.33} & 0.00 & 0.00 & 0.00 \\
\bottomrule
\end{tabular}%
}
\end{table}

\section{Method}
We propose an adaptive framework that improves instruction-guided image editing by reformulating the task and executing it in a structured, feedback-driven manner. This design improves alignment between task requirements and model behavior without modifying the underlying generative model.

\begin{figure*}[t]
    \centering
    \resizebox{\textwidth}{!}{%
        \includegraphics{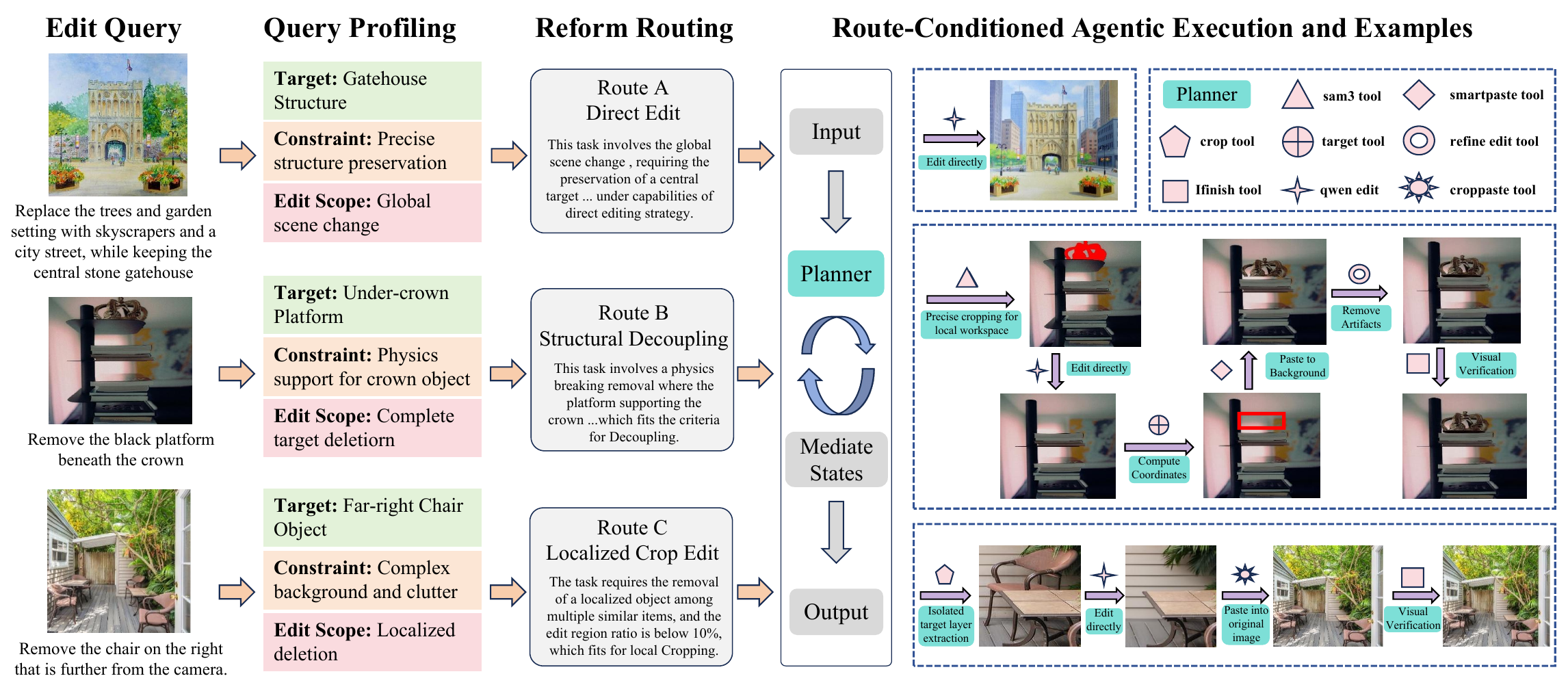}
    }
    \Description{A diagram illustrating the proposed adaptive task reformulation framework.}
    \caption{Overview of our framework. Each edit query is first profiled by its target, constraints, and scope, and then routed to a suitable reformulation strategy. A MLLM-based planner performs route-conditioned agentic execution with intermediate states and verification to obtain the final edited image.}
    \label{fig:frameff}
\end{figure*}

\subsection{Problem Formulation.}
Existing approaches typically formulate image editing as a direct mapping from an input image $I$ and a textual instruction $T$ to the target image $I^*$ given a speific editing model $M$, which can be formulated as:
\begin{equation*}
I^* = M(I, T).
\end{equation*}
This formulation assumes that the original input pair $(I, T)$ is already suitable for the model to execute the desired edit reliably. However, in many real cases, this assumption does not hold, especially When the target is small or the instruction is under-specified.

To address this issue, we reformulate image editing as a task reformulation problem. Given an input image--instruction pair $(I, T)$ and an editing model $M$, we first employ a multimodal model to get the edit query profile $p$, which captures target character and edit scopes. Based on $p$, the Reform Router $R$ determines an appropriate reformulation strategy $s$.
Conditioned on the selected strategy $s$, the agentic sequential inference is conducted over the tool set associated with the editing model, denoted as $\mathrm{Tools}(M)$. At each step, the agent selects and executes an operation grounded in the current state. This iterative process continues until the final edited image $I^*$ is obtained, as shown in Figure~\ref{fig:frameff}. The whole process can be formulated as:
\begin{equation*}
I^* = ATR(I, T, Tools(M)),
\end{equation*}
where $ATR$ is our method.

\subsection{Query Profile Awared Strategy Selection}

As shown by the pilot study in Sec. 3, the benefit of task reformulation is highly case-dependent: different failure subsets are recovered by different reformulation strategies, and no single transformation works uniformly well across all cases. This suggests that reformulation should not be applied in a fixed manner. Instead, the system must first infer what makes the current query difficult, and then select a suitable execution route accordingly.

To this end, we introduce a query profile 
$p$ as an intermediate representation for routing. Given an input image–instruction pair 
$(I, T)$, a edit profiler $P_{\text{profile}}$ first summarizes the editing query into a structured profile:
\[
p = {P}_{\text{profile}}(I, T),
\]
where profile $p$ captures the task characteristics that are most relevant to reformulation, including the edit target, the key constraint, and the effective edit scope. Concretely, it specifies which object or region should be manipulated, whether the edit involves ambiguity, structural dependency, or complex background coupling, and whether the requested modification is localized or scene-level. Compared with the raw input pair, this profile exposes the latent factors that determine whether direct execution is likely to fail.

Based on the observations from the pilot study, we organize reformulation routing into three execution routes:

\begin{equation*}
\begin{aligned}
S \in \{\, &\text{Route A: direct/rewrite execution}, \\
          &\text{Route B: spatial decoupling}, \\
          &\text{Route C: localized editing} \,\}.
\end{aligned}
\end{equation*}
Rather than treating all cases as equally distinct categories, we adopt a more practical decomposition. Specifically, we first identify whether a query requires localized editing or spatial decoupling, since these two cases consistently benefit from explicit input transformation in our pilot analysis. Queries that do not fall into either category are assigned to Route A, which covers the remaining cases that can still be handled in the original image space. Within Route A, we further determine whether the instruction is already sufficiently clear for direct editing, or whether it should first be rewritten into a more explicit and executable form before being passed to the editor.
Formally, the routing process is written as:
\[
s = {R}_{\text{router}}(I, T, p).
\]
where ${R}_{\text{router}}$ denotes the reformulation router. This design follows the empirical findings in Sec. 3: reformulation is most critical when the query exhibits strong locality or nontrivial spatial/structural dependency, while the remaining cases are better handled by lightweight semantic refinement rather than more aggressive scene decomposition. In this way, routing is driven not by a rigid taxonomy, but by the practical question of which form of task transformation is necessary to move the current query into a more reliable operating regime for the editing model.

\subsection{Route-Conditioned Agentic Execution}

Selecting a reformulation route only determines how the original editing task should be posed; it does not by itself specify a complete execution procedure. Even within the same route, different queries may require different tool sequences, intermediate transformations, and correction steps. Therefore, instead of using a fixed hand-crafted pipeline for each route, we perform route-conditioned agentic execution, where a planner dynamically decides the next operation based on the current intermediate state.

Formally, after obtaining the selected route 
$s$ , we define the initial execution state as
\[
s_0 = (I, T, S),
\]
where $I$ is the input image, $T$ is the instruction, and $S$ is the selected reformulation strategy.

During execution, the system maintains an intermediate state
\[
s_k = (I_k, T_k, S, H_k).
\]
where $I_k$ and $T_k$ denote the current image and instruction, respectively, and $H_k$
 denotes the execution history, including previously invoked tools and intermediate results. At each step, a planner generates the next action conditioned on both the selected route and the current state:
\[
a_{k+1} = \mathcal{T}_{\text{planner}}(s_k),
\]
where 
$a_{k+1}$
 may correspond to instruction rewriting, target localization, region cropping, object isolation, coordinate estimation, image editing, patch composition, artifact refinement, or termination. After executing the selected action, the environment updates the state as
 \[
s_{k+1} = \mathcal{E}(s_k, a_{k+1}),
\]
where $\mathcal{E}$ denotes the execution of the corresponding tool or operation. This process continues until the planner determines that the task has been satisfactorily completed, or a predefined step budget is reached.

\textbf{Direct Editing (Route A).}
This route is used for cases that do not require explicit spatial decoupling or localization. The planner operates in the original image space and decides whether the instruction can be executed directly or should first be rewritten into a more explicit and model-friendly form. When the instruction is already clear and well grounded, the editor is applied directly. Otherwise, the planner reformulates the instruction to reduce ambiguity, clarify the target, or make the desired transformation more explicit, and then performs editing on the full image.

\textbf{Spatial Decoupling (Pipeline B).}
This route is designed for queries involving structural dependency, object support, explicit relocation, or strong coupling between the target and surrounding content. In such cases, direct editing often leads to broken geometry or implausible composition. The planner therefore decomposes the task into structured operations, including target isolation, destination estimation or structure-aware manipulation, background completion over the original region, and recomposition of the edited content into the scene. Optional refinement is further applied to suppress boundary artifacts and improve visual consistency. By separating manipulation from scene reconstruction, this route converts a difficult global edit into more controllable subproblems.

\textbf{Localized Editing (Pipeline C).}
This route is used when the target is small, weakly grounded, or embedded in a cluttered scene. Instead of editing the full image directly, the planner first estimates a target-centered local workspace and performs editing in the cropped region, where the target occupies a larger effective proportion of the input. The edited patch is then pasted back into the original image, optionally followed by local refinement to improve boundary smoothness and contextual consistency. This route improves the signal-to-noise ratio of the editing problem and is particularly effective for small-object manipulation and fine-grained target selection.

A key property of our framework is that execution is closed-loop rather than one-shot. Intermediate outputs are continuously inspected to determine whether the current result satisfies the route-specific objective or whether additional correction steps are needed. For example, the planner may trigger refinement after imperfect composition, re-localize the target when cropping is inaccurate, or regenerate an edit when the rewritten instruction still fails. This feedback-driven design improves robustness under long or ambiguous editing queries, and its effectiveness is validated by the ablation results in Table~\ref{tab:router_agentic_ablation}.

To further stabilize inference, we incorporate a bounded fallback mechanism. If execution fails to meet the termination condition within a fixed number of steps, the system falls back to a safe single-pass edit on the original input. This limits error accumulation while preserving practical efficiency. As shown in Table~\ref{tab:router_agentic_ablation}, the fallback design also yields additional gains, confirming its role in improving the reliability of multi-step execution. Overall, route-conditioned agentic execution enables ATR to unify structured reformulation, dynamic tool composition, and intermediate feedback within a single inference-time framework.

\section{Experiments}

\begin{table*}[t]
\centering
\caption{Quantitative comparison on ImgEdit-Easy and ImgEdit-Hard benchmarks.}
\label{tab:imgedit_main}
\setlength{\tabcolsep}{4.5pt}
\renewcommand{\arraystretch}{1.1}
\resizebox{\textwidth}{!}{%
\begin{tabular}{lccccccccccc}
\toprule
\multirow{2}{*}{Method}
& \multicolumn{10}{c}{ImgEdit-Easy}
& \multicolumn{1}{c}{ImgEdit-Hard} \\
\cmidrule(lr){2-11} \cmidrule(lr){12-12}
& Add $\uparrow$ & Background $\uparrow$ & Extract $\uparrow$ & Remove $\uparrow$ & Replace $\uparrow$ & Style $\uparrow$ & Action $\uparrow$ & Adjust $\uparrow$ & Compose $\uparrow$ & Avg. $\uparrow$ & Avg. $\uparrow$ \\
\midrule

\multicolumn{12}{l}{\textit{Open-source baselines}} \\
Qwen-Edit             & 4.20 & 3.98 & 2.23 & 3.96 & 4.36 & 4.84 & \textbf{4.48} & 4.22 & 3.07 & 3.93 & 3.57 \\
RePlan                & 3.59 & 3.55 & 1.52 & 3.38 & 3.71 & 4.73 & 3.02 & 3.52 & 2.99 & 3.33 & 3.52 \\
EditThinker           & 4.14 & 4.02 & 2.88 & 4.01 & 4.20 & 4.60 & 4.28 & 4.12 & 3.40 & 3.96 & 3.60 \\

\midrule
\multicolumn{12}{l}{\textit{Closed-source baselines}} \\
Nano Banana           & 4.44 & 4.13 & 2.65 & 4.14 & 4.51 & 3.75 & 4.16 & 4.46 & 3.54 & 3.98 & 3.88 \\
Nano Banana Pro       & \textbf{4.49} & 4.19 & 2.37 & \underline{4.46} & 4.42 & \textbf{4.87} & 4.37 & \textbf{4.59} & \textbf{3.90} & \underline{4.18} & \textbf{4.25} \\

\midrule
\multicolumn{12}{l}{\textit{Our variants}} \\
Qwen-Edit-Bo2         & 4.22 & 4.03 & 2.30 & 4.04 & 4.42 & \underline{4.85} & 4.37 & 4.17 & 3.27 & 3.96 & 3.81 \\
Qwen-Edit-ATR         & 4.16 & 4.01 & \textbf{3.29} & 3.86 & 4.45 & 4.72 & \underline{4.43} & 4.24 & \underline{3.83} & 4.11 & 4.13 \\
Nano Banana-Bo2       & \underline{4.47} & \textbf{4.50} & 2.85 & 4.42 & \underline{4.53} & 4.10 & 4.29 & 4.45 & 3.58 & 4.13 & 4.01 \\
Nano Banana-ATR       & \underline{4.47} & \underline{4.23} & \underline{3.28} & \textbf{4.48} & \textbf{4.63} & 4.03 & \underline{4.43} & \underline{4.53} & 3.72 & \textbf{4.20} & \underline{4.19} \\

\bottomrule
\end{tabular}%
}
\end{table*}
\begin{figure*}[t]
    \centering
    \resizebox{0.95\textwidth}{!}{%
        \includegraphics{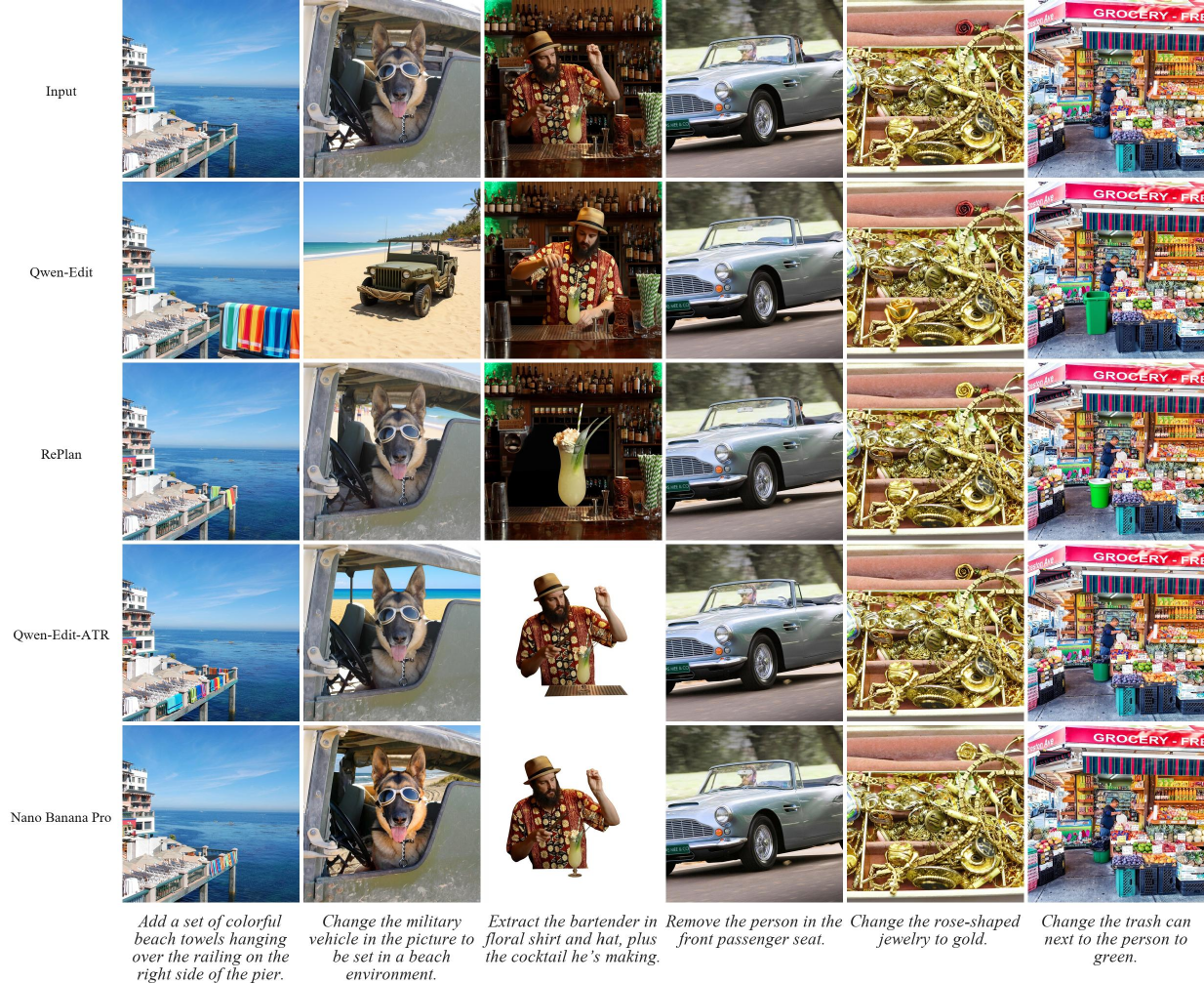}
    }
    \Description{A grid of images showing qualitative editing comparisons on the ImgEdit benchmark.}
    \caption{Qualitative editing results on the ImgEdit benchmark. Compared to direct editing baselines, our ATR framework demonstrates superior performance in fine-grained target isolation, precise spatial grounding, and attribute editing while preserving background consistency.}
    \label{fig:imgedit_qualitative}
\end{figure*}

\subsection{Experimental Setup}

\textbf{Benchmarks.}
We evaluate our method on three complementary benchmarks: \textbf{ImgEdit-Bench} serves as a standard testbed for general-purpose editing across diverse categories (e.g., addition, removal, and attribute modification). \textbf{PICA} is a comprehensive and fine-grained benchmark dedicated to physically realistic image editing. It utilizes a region-aware, VQA-based evaluation protocol to provide interpretable and reliable assessments of physical realism, ensuring plausible scene integration and global coherence. \textbf{RePlan (IV-Edit)} emphasizes instruction--visual complexity in cluttered scenes, stressing precise grounding and localization of ambiguous instructions among multiple targets. Together, they cover general editing capability, physical plausibility, and fine-grained grounding.

\textbf{Baselines.}
We compare our method against three representative paradigms: (1) \textbf{Direct models} (\textit{Qwen-Edit}, \textit{Nano Banana/Pro}), which follow the standard direct-mapping paradigm but lack explicit intermediate reasoning or structural decomposition. (2) \textbf{Reasoning-augmented frameworks} (\textit{EditThinker}), which utilizes an iterative \emph{think-while-edit} loop for feedback-driven refinement, though it lacks structured control over spatial operations. (3) \textbf{Planning-based methods} (\textit{RePlan}), which decomposes instructions into region-aligned guidance via structured planning but relies on a fixed, non-adaptive formulation.

\textbf{Controlled variants and implementation.}
We evaluate \textit{Qwen-Edit} and \textit{Nano Banana} under two controlled settings: \textit{Bo2} (Best-of-2) and our \textit{ATR}. The \textit{Bo2} variants generate two candidates and select the best, a strategy explicitly designed to align the inference cost with our multi-step framework for fair comparison. Specifically, the average number of backbone calls for \textit{Qwen-Edit-ATR-Edit} is consistently between 1 and 2 per task (1.35 on ImgEdit-Basic, 1.71 on ImgEdit-Hard, 1.2 on PICA, and 1.5 on RePlan). Finally, all reported quantitative scores are computed by averaging the results of three independent evaluation runs to mitigate generative variance.

Overall, these baselines span from direct generation to iterative reasoning and explicit planning. This enables a comprehensive evaluation to verify that the improvements of our method arise from adaptive input reformulation and agentic decision-making at inference time, rather than from backbone model differences alone.

\subsection{Main Results}
We evaluate our method on ImgEdit-Bench, PICA, and RePlan, comparing against representative baselines and computationally intensive variants. These experiments aim to comprehensively assess the effectiveness of our approach under different levels of task complexity. The detailed results are presented below:

\subsubsection{ImgEdit Bench}

As shown in Table~\ref{tab:imgedit_main}, ATR consistently improves performance across different base models by addressing specific execution failures. Quantitatively, by reformulating ambiguous instructions into sub-tasks with explicit local constraints, it improves Qwen-Edit's performance on the \textit{Extract} and \textit{Compose} tasks by 1.06 and 0.76, respectively. This method also benefits lightweight models: Nano Banana achieves 4.20 on ImgEdit-Easy, outperforming the Nano Banana Pro baseline (4.18). Furthermore, ATR consistently surpasses the Best-of-2 strategy, indicating gains beyond passive multi-sampling. On the challenging ImgEdit-Hard benchmark, which features complex and multi-target instructions, Qwen-Edit-ATR achieves a score of 4.13, significantly exceeding its base model (3.57) and the planning-based RePlan (3.52).

These quantitative improvements are visually corroborated in Figure~\ref{fig:imgedit_qualitative}. Direct editing models often struggle with precise target localization and scene preservation. For instance, given a complex spatial instruction (Column 1), Qwen-Edit completely fails to locate the designated railing on the pier to ground the towels, resulting in an editing failure where it hallucinated a different scene altogether. In contrast, Nano Banana Pro achieves the optimal result, closely followed by Qwen-Edit-ATR, which successfully adds the colorful towels while preserving the original background with minimal disruption. Furthermore, our method demonstrates high precision in fine-grained target isolation and attribute editing. It accurately identifies and removes the passenger (Column 4). Similarly, when altering the trash can (Column 5), other baselines inadvertently mutate the object's structural shape while altering its color, whereas our method and Nano Banana Pro strictly manipulate the specified color attribute while preserving the original morphology.

\begin{table}[t]
\centering
\caption{Quantitative comparison on PICA and RePlan benchmarks.}
\label{tab:main_results}
\renewcommand{\arraystretch}{1.1}
\resizebox{\columnwidth}{!}{
\begin{tabular}{l c ccccc}
\toprule
\multirow{2}{*}{Method} 
& \multicolumn{1}{c}{PICA} 
& \multicolumn{5}{c}{RePlan} \\
\cmidrule(lr){2-2} \cmidrule(lr){3-7}
& Avg. $\uparrow$
& Qua. $\uparrow$ & Target. $\uparrow$ & Effect $\uparrow$ & Consist. $\uparrow$ & Avg. $\uparrow$ \\
\midrule

\multicolumn{7}{l}{\textit{Open-source baselines}} \\
Qwen-Edit            & 61.43\% & 3.62 & 3.85 & 3.53 & 2.39 & 3.35 \\
RePlan               & 50.18\% & 3.82 & 3.79 & 3.05 & 3.51 & 3.54 \\
EditThinker          & 62.17\% & 3.74 & 3.84 & 3.92 & 2.55 & 3.51 \\
\midrule
\multicolumn{7}{l}{\textit{Closed-source baselines}} \\
Nano Banana          & 60.73\% & 4.01 & 4.28 & 3.97 & 2.90 & 3.79 \\
Nano Banana Pro      & \textbf{66.14\%} & 4.09 & \textbf{4.75} & \textbf{4.69} & 3.56 & \textbf{4.27} \\
\midrule
\multicolumn{7}{l}{\textit{Our variants}} \\
Qwen-Edit-Bo2        & 63.68\% & 3.64 & 4.02 & 3.57 & 2.38 & 3.40 \\
Qwen-Edit-ATR        & \underline{65.91\%} & 3.86 & \underline{4.58} & \underline{4.39} & \underline{3.72} & 4.14 \\
Nano Banana-Bo2      & 61.75\% & \textbf{4.12} & 4.42 & 4.15 & 3.23 & 3.98 \\
Nano Banana-ATR      & 63.45\% & \textbf{4.12} & 4.47 & 4.32 & \textbf{4.00} & \underline{4.23} \\

\bottomrule
\end{tabular}
}
\end{table}

\subsubsection{PICA and RePlan Benchmarks}

Table~\ref{tab:main_results} evaluates the models on complex spatial mapping and physical consistency. On the RePlan benchmark, which emphasizes instruction-visual complexity in cluttered scenes, direct editing models often suffer from localization shifts. For instance, Qwen-Edit achieves a consistency score of only 2.39. By decoupling complex instructions into local execution sequences via the agent, ATR eliminates referential ambiguity and precise region alignment, boosting the overall score to 4.14. Furthermore, compared to the multi-sampling strategy which yields negligible improvements here, ATR demonstrates that active reformulation provides gains beyond passive trial-and-error. 

Since EditThinker's default prompt struggles on the RePlan dataset, risking an underestimation of its capabilities, we tailored its prompt specifically for this evaluation to ensure fairness. Its original configuration remains strictly unchanged across all other benchmarks.

\begin{figure}[t]
    \centering
    \resizebox{\columnwidth}{!}{%
        \includegraphics{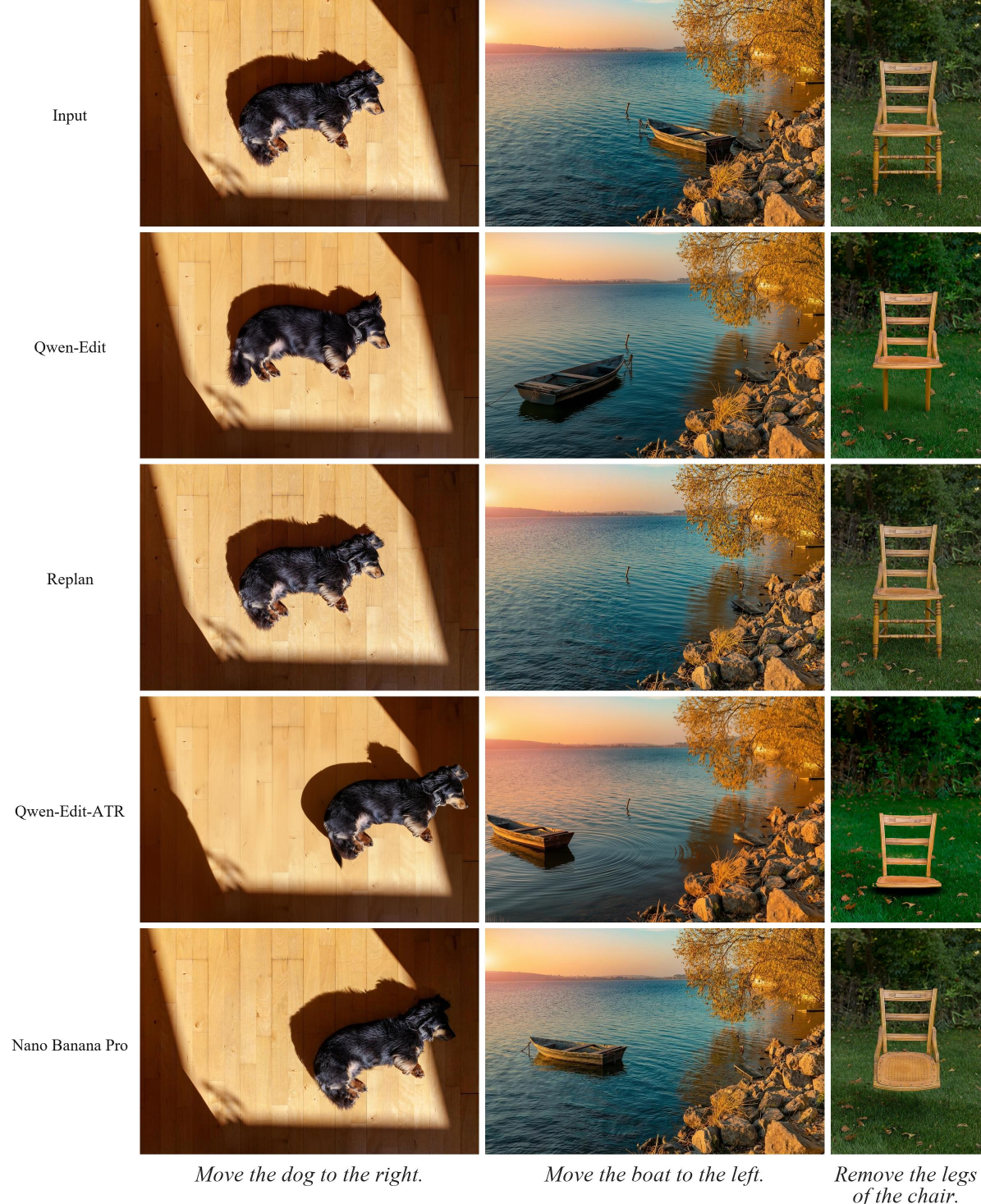}
    }
    \Description{Visual comparisons of editing results on the PICA benchmark showing precise spatial relocation.}
    \caption{Qualitative editing results on the PICA benchmark. Our method effectively bypasses failure modes such as spatial translation errors and target disappearance, achieving accurate relocation and pixel-level structural decoupling.}
    \label{fig:pica_qualitative}
\end{figure}

On the PICA benchmark, which evaluates global geometry and physical consistency, equipping base models with our ATR framework provides a substantial boost—elevating Qwen-Edit from 61.43\% to 65.91\% and Nano Banana from 60.73\% to 63.45\%. These quantitative gains align with the visual results in Figure~\ref{fig:pica_qualitative}. In spatial translation and structural decoupling tasks, traditional models frequently fail by ignoring movement directives (e.g., the dog, left column), inadvertently erasing targets (e.g., the boat, middle column), or excessively erasing entire objects under local constraints (e.g., removing chair legs). In contrast, ATR ensures accurate relocation and utilizes SAM3 for pixel-level structural decoupling, effectively preserving unmentioned components (e.g., the seat cushion). Furthermore, this agentic execution effectively compensates for limitations in model capacity: the base Nano Banana-ATR closely approaches the performance of its much larger counterpart, Nano Banana Pro. This demonstrates that task reformulation serves as an efficient alternative to brute-force model scaling.

\subsection{Ablation Studies}

To evaluate the contribution of each component within ATR, we conduct an ablation study on the ImgEdit-Hard benchmark (Table~\ref{tab:reformulation_ablation}).

\textbf{Incremental Expansion of Routing Decision Space.} 
The baseline (config a), limited to direct execution, scores 3.57. Introducing the \textit{Rewrite} module (config b) enhances semantic clarity, raising the score to 3.82. Further incorporating SAM-based \textit{Spatial} reformulation (config c) aids precise geometric transformations, yielding 3.87. Adding a \textit{Local} processing path (config d) enables target isolation and reduces visual complexity, boosting performance to 4.02. Finally, conditioning the routing decisions on global \textit{Context} (config e) achieves the optimal score of 4.16. These results demonstrate that a context-aware routing logic can precisely select the most effective strategy, balancing local editing fidelity with global scene consistency.

\textbf{Robustness of the Agentic Execution Pipeline.} 
Table~\ref{tab:router_agentic_ablation} evaluates the contribution of each agentic execution component. Direct editing in configuration $a$ scores 3.57. Relying solely on automated routing in configuration $b$ or oracle routing in configuration $c$ yields limited scores of 3.62 and 3.70, respectively. Integrating MDP-based constraints in configuration $d$ structures the execution logic and reaches 3.74, while the fallback mechanism in configuration $e$ improves error recovery, raising the score to 3.89. The addition of terminal verification (\textit{If-finish}) in configuration $f$ establishes a closed-loop system for iterative refinement, significantly boosting the score to 4.16. Furthermore, the minimal performance gap between the full automated pipeline in configuration $f$ and the oracle-guided pipeline in configuration $g$, which scores 4.22, demonstrates the strong robustness of our framework.
\subsection{Supplementary Material}

Due to space constraints, the supplementary material provides extended evaluations and technical details. Specifically, \textbf{Appendix A} presents additional qualitative comparisons across benchmarks. \textbf{Appendix B} showcases step-by-step visual case studies of agentic execution. \textbf{Appendix C} details the ImgEdit pilot study and failure analysis. To facilitate reproducibility, \textbf{Appendix D} documents all underlying tool implementations. \textbf{Appendix E} analyzes limitations regarding conflicts between localized edits and global attributes. Finally, \textbf{Appendix F} details the refinement of the PICA evaluation protocol to ensure a fair assessment of physical realism.
\subsection{Limitations and Future Work}

Although our ATR framework demonstrates significant advantages over existing baselines by utilizing adaptive task reformulation to handle difficult editing cases requiring fine-grained control and local awareness, complex image editing still presents highly challenging corner cases. 

Specifically, when an editing instruction relies heavily on both precise local anchoring and complex global information coordination, existing visual generative models often struggle to perfectly balance the two. While our framework mitigates this issue to a certain extent, there remains room for optimization. We provide a detailed analysis of such extreme failure cases in the supplementary material. Moving forward, exploring how to more effectively decouple and synergize local features with global contexts in future agent routing and reformulation strategies---thereby overcoming the inherent bottlenecks of base models---will be a primary focus of our future work.

\begin{table}[t]
\centering
\caption{Ablation configurations of reformulation components.}
\label{tab:reformulation_ablation}
\renewcommand{\arraystretch}{1.1}
\resizebox{\columnwidth}{!}{
\begin{tabular}{ccccccc}
\toprule
\multirow{2}{*}{Config} 
& \multicolumn{4}{c}{Reformulation Task} 
& \multirow{2}{*}{Context} 
& \multirow{2}{*}{Avg score $\uparrow$} \\
\cmidrule(lr){2-5}
& Direct & Rewrite & Spatial & Local &  &  \\
\midrule
a & \checkmark &  &  &  &  & 3.57 \\
b & \checkmark & \checkmark &  &  &  & 3.82 \\
c & \checkmark & \checkmark & \checkmark &  &  & 3.87 \\
d & \checkmark & \checkmark & \checkmark & \checkmark &  & \underline{4.02} \\
e & \checkmark & \checkmark & \checkmark & \checkmark & \checkmark & \textbf{4.16} \\
\bottomrule
\end{tabular}
}
\end{table}

\begin{table}[t]
\centering
\caption{Ablation of routing and agentic execution components.}
\label{tab:router_agentic_ablation}
\renewcommand{\arraystretch}{1.1}
\resizebox{\columnwidth}{!}{
\begin{tabular}{lcccccc}
\toprule
\multirow{2}{*}{Config} 
& \multirow{2}{*}{Router} 
& \multirow{2}{*}{Oracle} 
& \multicolumn{3}{c}{Agentic} 
& \multirow{2}{*}{Avg score $\uparrow$} \\
\cmidrule(lr){4-6}
&  &  & MDP & Fallback & If-finish &  \\
\midrule
a &  &  &  &  &  & 3.57 \\
b & \checkmark &  &  &  &  & 3.62 \\
c &  & \checkmark &  &  &  & 3.70 \\
d & \checkmark &  & \checkmark &  &  & 3.74 \\
e & \checkmark &  & \checkmark & \checkmark &  & 3.89 \\
f & \checkmark &  & \checkmark & \checkmark & \checkmark & \underline{4.16} \\
g &  & \checkmark & \checkmark & \checkmark & \checkmark & \textbf{4.22} \\
\bottomrule
\end{tabular}
}
\end{table}

\section{Conclusion}

In this work, we revisit instruction-guided image editing from the perspective of \emph{task formulation mismatch}. We argue that many failures arise not because the model lacks the required capability, but because the original image--instruction pair is poorly conditioned for reliable execution. Based on this observation, we propose ATR, an adaptive framework that improves editing reliability through query profiling, reformulation routing, and route-conditioned agentic execution, without modifying the backbone editor.
Experiments on ImgEdit, PICA, and RePlan across multiple editing backbones show consistent improvements, especially on challenging cases involving ambiguity, small targets, and complex spatial dependency. These results suggest that task reformulation is a critical yet underexplored dimension of image editing, complementary to advances in model architecture and training.



\bibliographystyle{ACM-Reference-Format}
\bibliography{sample-base}

@String{Computing = "Computing" }

@String{Computer = "{IEEE} Computer" }

@inproceedings{brooks2023instructpix2pix,
  title={Instructpix2pix: Learning to follow image editing instructions},
  author={Brooks, Tim and Holynski, Aleksander and Efros, Alexei A},
  booktitle={Proceedings of the IEEE/CVF Conference on Computer Vision and Pattern Recognition},
  pages={18392--18402},
  year={2023}
}

@inproceedings{zhang2023magicbrush,
  title={Magicbrush: A manually annotated dataset for instruction-guided image editing},
  author={Zhang, Kai and Mo, Lingbo and Chen, Wenhu and Sun, Huan and Su, Yu},
  booktitle={Advances in Neural Information Processing Systems},
  volume={36},
  pages={31428--31449},
  year={2023}
}

@inproceedings{zhang2023adding,
  title={Adding conditional control to text-to-image diffusion models},
  author={Zhang, Lvmin and Rao, Anyi and Agrawala, Maneesh},
  booktitle={Proceedings of the IEEE/CVF international conference on computer vision},
  pages={3836--3847},
  year={2023}
}

@inproceedings{sheynin2024emu,
  title={Emu edit: Precise image editing via recognition and generation tasks},
  author={Sheynin, Shelly and Polyak, Adam and Singer, Uriel and Kirstain, Yuval and Zohar, Amit and Ashual, Oron and Parikh, Devi and Taigman, Yaniv},
  booktitle={Proceedings of the IEEE/CVF Conference on Computer Vision and Pattern Recognition},
  pages={8871--8879},
  year={2024}
}

@inproceedings{huang2024smartedit,
  title={Smartedit: Exploring complex instruction-based image editing with multimodal large language models},
  author={Huang, Yuzhou and Xie, Liangbin and Wang, Xintao and Yuan, Ziyang and Cun, Xiaodong and Ge, Yixiao and Zhou, Jiantao and Dong, Chao and Huang, Rui and Zhang, Ruimao and others},
  booktitle={Proceedings of the IEEE/CVF Conference on Computer Vision and Pattern Recognition},
  pages={8362--8371},
  year={2024}
}

@article{shen2023hugginggpt,
  title={Hugginggpt: Solving ai tasks with chatgpt and its friends in hugging face},
  author={Shen, Yongliang and Song, Kaitao and Tan, Xu and Li, Dongsheng and Lu, Weiming and Zhuang, Yueting},
  journal={Advances in Neural Information Processing Systems},
  volume={36},
  pages={38154--38180},
  year={2023}
}

@article{pu2025picabench,
  title={PICABench: How Far Are We from Physically Realistic Image Editing?},
  author={Pu, Yuandong and Zhuo, Le and Han, Songhao and Xing, Jinbo and Zhu, Kaiwen and Cao, Shuo and Fu, Bin and Liu, Si and Li, Hongsheng and Qiao, Yu and others},
  journal={arXiv preprint arXiv:2510.17681},
  year={2025}
}

@article{qu2025replan,
  title={RePlan: Reasoning-guided Region Planning for Complex Instruction-based Image Editing},
  author={Qu, Tianyuan and Ke, Lei and Zhan, Xiaohang and Tang, Longxiang and Liu, Yuqi and Peng, Bohao and Yu, Bei and Yu, Dong and Jia, Jiaya},
  journal={arXiv preprint arXiv:2512.16864},
  year={2025}
}

@article{ye2025imgedit,
  title={Imgedit: A unified image editing dataset and benchmark},
  author={Ye, Yang and He, Xianyi and Li, Zongjian and Lin, Bin and Yuan, Shenghai and Yan, Zhiyuan and Hou, Bohan and Yuan, Li},
  journal={arXiv preprint arXiv:2505.20275},
  year={2025}
}

@inproceedings{esser2024scaling,
  title={Scaling rectified flow transformers for high-resolution image synthesis},
  author={Esser, Patrick and Kulal, Sumith and Blattmann, Andreas and Entezari, Rahim and M{\"u}ller, Jonas and Saini, Harry and Levi, Yam and Lorenz, Dominik and Sauer, Axel and Boesel, Frederic and others},
  booktitle={Forty-first international conference on machine learning},
  year={2024}
}

@article{wu2025qwen,
  title={Qwen-image technical report},
  author={Wu, Chenfei and Li, Jiahao and Zhou, Jingren and Lin, Junyang and Gao, Kaiyuan and Yan, Kun and Yin, Sheng-ming and Bai, Shuai and Xu, Xiao and Chen, Yilei and others},
  journal={arXiv preprint arXiv:2508.02324},
  year={2025}
}

@article{basu2023editval,
  title={Editval: Benchmarking diffusion based text-guided image editing methods},
  author={Basu, Samyadeep and Saberi, Mehrdad and Bhardwaj, Shweta and Chegini, Atoosa Malemir and Massiceti, Daniela and Sanjabi, Maziar and Hu, Shell Xu and Feizi, Soheil},
  journal={arXiv preprint arXiv:2310.02426},
  year={2023}
}

@article{zhao2025learning,
  title={Learning position-aware implicit neural network for real-world face inpainting},
  author={Zhao, Bo and Yang, Huan and Fu, Jianlong},
  journal={Pattern Recognition},
  volume={165},
  pages={111598},
  year={2025},
  publisher={Elsevier}
}

@article{li2025editthinker,
  title={Editthinker: Unlocking iterative reasoning for any image editor},
  author={Li, Hongyu and Zhang, Manyuan and Zheng, Dian and Guo, Ziyu and Jia, Yimeng and Feng, Kaituo and Yu, Hao and Liu, Yexin and Feng, Yan and Pei, Peng and others},
  journal={arXiv preprint arXiv:2512.05965},
  year={2025}
}

@article{wang2024genartist,
  title={Genartist: Multimodal llm as an agent for unified image generation and editing},
  author={Wang, Zhenyu and Li, Aoxue and Li, Zhenguo and Liu, Xihui},
  journal={Advances in Neural Information Processing Systems},
  volume={37},
  pages={128374--128395},
  year={2024}
}

@inproceedings{kawar2023imagic,
  title={Imagic: Text-based real image editing with diffusion models},
  author={Kawar, Bahjat and Zada, Shiran and Lang, Oran and Tov, Omer and Chang, Huiwen and Dekel, Tali and Mosseri, Inbar and Irani, Michal},
  booktitle={Proceedings of the IEEE/CVF conference on computer vision and pattern recognition},
  pages={6007--6017},
  year={2023}
}

@article{fu2023guiding,
  title={Guiding instruction-based image editing via multimodal large language models},
  author={Fu, Tsu-Jui and Hu, Wenze and Du, Xianzhi and Wang, William Yang and Yang, Yinfei and Gan, Zhe},
  journal={arXiv preprint arXiv:2309.17102},
  year={2023}
}

@inproceedings{zhang2023sine,
  title={Sine: Single image editing with text-to-image diffusion models},
  author={Zhang, Zhixing and Han, Ligong and Ghosh, Arnab and Metaxas, Dimitris N and Ren, Jian},
  booktitle={Proceedings of the IEEE/CVF conference on computer vision and pattern recognition},
  pages={6027--6037},
  year={2023}
}

@article{podell2023sdxl,
  title={Sdxl: Improving latent diffusion models for high-resolution image synthesis},
  author={Podell, Dustin and English, Zion and Lacey, Kyle and Blattmann, Andreas and Dockhorn, Tim and M{\"u}ller, Jonas and Penna, Joe and Rombach, Robin},
  journal={arXiv preprint arXiv:2307.01952},
  year={2023}
}

@article{betker2023improving,
  title={Improving image generation with better captions},
  author={Betker, James and Goh, Gabriel and Jing, Li and Brooks, Tim and Wang, Jianfeng and Li, Linjie and Ouyang, Long and Zhuang, Juntang and Lee, Joyce and Guo, Yufei and others},
  journal={Computer Science. https://cdn. openai. com/papers/dall-e-3. pdf},
  volume={2},
  number={3},
  pages={8},
  year={2023}
}

@inproceedings{geng2024instructdiffusion,
  title={Instructdiffusion: A generalist modeling interface for vision tasks},
  author={Geng, Zigang and Yang, Binxin and Hang, Tiankai and Li, Chen and Gu, Shuyang and Zhang, Ting and Bao, Jianmin and Zhang, Zheng and Li, Houqiang and Hu, Han and others},
  booktitle={Proceedings of the IEEE/CVF Conference on computer vision and pattern recognition},
  pages={12709--12720},
  year={2024}
}

@article{hao2023optimizing,
  title={Optimizing prompts for text-to-image generation},
  author={Hao, Yaru and Chi, Zewen and Dong, Li and Wei, Furu},
  journal={Advances in Neural Information Processing Systems},
  volume={36},
  pages={66923--66939},
  year={2023}
}

@article{lian2023llm,
  title={Llm-grounded diffusion: Enhancing prompt understanding of text-to-image diffusion models with large language models},
  author={Lian, Long and Li, Boyi and Yala, Adam and Darrell, Trevor},
  journal={arXiv preprint arXiv:2305.13655},
  year={2023}
}

@inproceedings{liu2022design,
  title={Design guidelines for prompt engineering text-to-image generative models},
  author={Liu, Vivian and Chilton, Lydia B},
  booktitle={Proceedings of the 2022 CHI conference on human factors in computing systems},
  pages={1--23},
  year={2022}
}

@article{zhao2026texeditor,
  title={TexEditor: Structure-Preserving Text-Driven Texture Editing},
  author={Zhao, Bo and Liu, Yihang and Zhang, Chenfeng and Yang, Huan and Gai, Kun and Ji, Wei},
  journal={arXiv preprint arXiv:2603.18488},
  year={2026}
}

@inproceedings{cao2023beautifulprompt,
  title={Beautifulprompt: Towards automatic prompt engineering for text-to-image synthesis},
  author={Cao, Tingfeng and Wang, Chengyu and Liu, Bingyan and Wu, Ziheng and Zhu, Jinhui and Huang, Jun},
  booktitle={Proceedings of the 2023 Conference on Empirical Methods in Natural Language Processing: Industry Track},
  pages={1--11},
  year={2023}
}

@inproceedings{yang2024mastering,
  title={Mastering Text-to-Image Diffusion: Recaptioning, Planning, and Generating with Multimodal LLMs.},
  author={Yang, Ling and Yu, Zhaochen and Meng, Chenlin and Xu, Minkai and Ermon, Stefano and Cui, Bin},
  booktitle={Icml},
  volume={3},
  number={6},
  pages={7},
  year={2024}
}

@inproceedings{pan2023drag,
  title={Drag your gan: Interactive point-based manipulation on the generative image manifold},
  author={Pan, Xingang and Tewari, Ayush and Leimk{\"u}hler, Thomas and Liu, Lingjie and Meka, Abhimitra and Theobalt, Christian},
  booktitle={ACM SIGGRAPH 2023 conference proceedings},
  pages={1--11},
  year={2023}
}

@inproceedings{ruiz2023dreambooth,
  title={Dreambooth: Fine tuning text-to-image diffusion models for subject-driven generation},
  author={Ruiz, Nataniel and Li, Yuanzhen and Jampani, Varun and Pritch, Yael and Rubinstein, Michael and Aberman, Kfir},
  booktitle={Proceedings of the IEEE/CVF conference on computer vision and pattern recognition},
  pages={22500--22510},
  year={2023}
}

@article{couairon2022diffedit,
  title={Diffedit: Diffusion-based semantic image editing with mask guidance},
  author={Couairon, Guillaume and Verbeek, Jakob and Schwenk, Holger and Cord, Matthieu},
  journal={arXiv preprint arXiv:2210.11427},
  year={2022}
}

@inproceedings{zhou2025fireedit,
  title={Fireedit: Fine-grained instruction-based image editing via region-aware vision language model},
  author={Zhou, Jun and Li, Jiahao and Xu, Zunnan and Li, Hanhui and Cheng, Yiji and Hong, Fa-Ting and Lin, Qin and Lu, Qinglin and Liang, Xiaodan},
  booktitle={Proceedings of the Computer Vision and Pattern Recognition Conference},
  pages={13093--13103},
  year={2025}
}

@article{huang2023t2i,
  title={T2i-compbench: A comprehensive benchmark for open-world compositional text-to-image generation},
  author={Huang, Kaiyi and Sun, Kaiyue and Xie, Enze and Li, Zhenguo and Liu, Xihui},
  journal={Advances in Neural Information Processing Systems},
  volume={36},
  pages={78723--78747},
  year={2023}
}

@inproceedings{wang2025sega,
  title={Sega: A stepwise evolution paradigm for content-aware layout generation with design prior},
  author={Wang, Haoran and Zhao, Bo and Wang, Jinghui and Wang, Hanzhang and Yang, Huan and Ji, Wei and Liu, Hao and Xiao, Xinyan},
  booktitle={Proceedings of the IEEE/CVF International Conference on Computer Vision},
  pages={19321--19330},
  year={2025}
}

@article{chen2025regione,
  title={RegionE: Adaptive Region-Aware Generation for Efficient Image Editing},
  author={Chen, Pengtao and Zeng, Xianfang and Zhao, Maosen and Shen, Mingzhu and Ye, Peng and Xiang, Bangyin and Wang, Zhibo and Cheng, Wei and Yu, Gang and Chen, Tao},
  journal={arXiv preprint arXiv:2510.25590},
  year={2025}
}

@inproceedings{tumanyan2023plug,
  title={Plug-and-play diffusion features for text-driven image-to-image translation},
  author={Tumanyan, Narek and Geyer, Michal and Bagon, Shai and Dekel, Tali},
  booktitle={Proceedings of the IEEE/CVF conference on computer vision and pattern recognition},
  pages={1921--1930},
  year={2023}
}

\clearpage
\appendix
\begin{center}
    \Large \textbf{Supplementary Material}
\end{center}

This appendix provides extended evaluations and technical details to support the findings presented in the main text. Section~\ref{sec:appendix_qualitative} showcases additional qualitative comparisons across the PICA, Replan, and ImgEdit benchmarks to further demonstrate the generalization of our framework. Section~\ref{sec:appendix_case_studies} provides step-by-step visual case studies of our agent's dynamic execution and toolchain invocations. Section~\ref{sec:appendix_pilot_imgedit} presents the detailed pilot study and failure-case analysis on the ImgEdit benchmark. To facilitate reproducibility, Section~\ref{sec:appendix_tools} rigorously documents the definitions, inputs/outputs, and core mechanisms of the underlying tools utilized in our pipeline. Section~\ref{sec:appendix_limitations} offers an in-depth limitation analysis, exploring extreme scenarios where localized agentic execution conflicts with global image attributes. Finally, Section~\ref{sec:appendix_pica_refinement} details the refinement of the PICA evaluation protocol, outlining the removal of ill-posed queries to ensure a rigorous and fair assessment of physical realism.

\section{Additional Qualitative Results}
\label{sec:appendix_qualitative}
\subsection{Results on the PICA Benchmark}

\begin{figure}[htbp]
    \centering
    \includegraphics[width=\columnwidth]{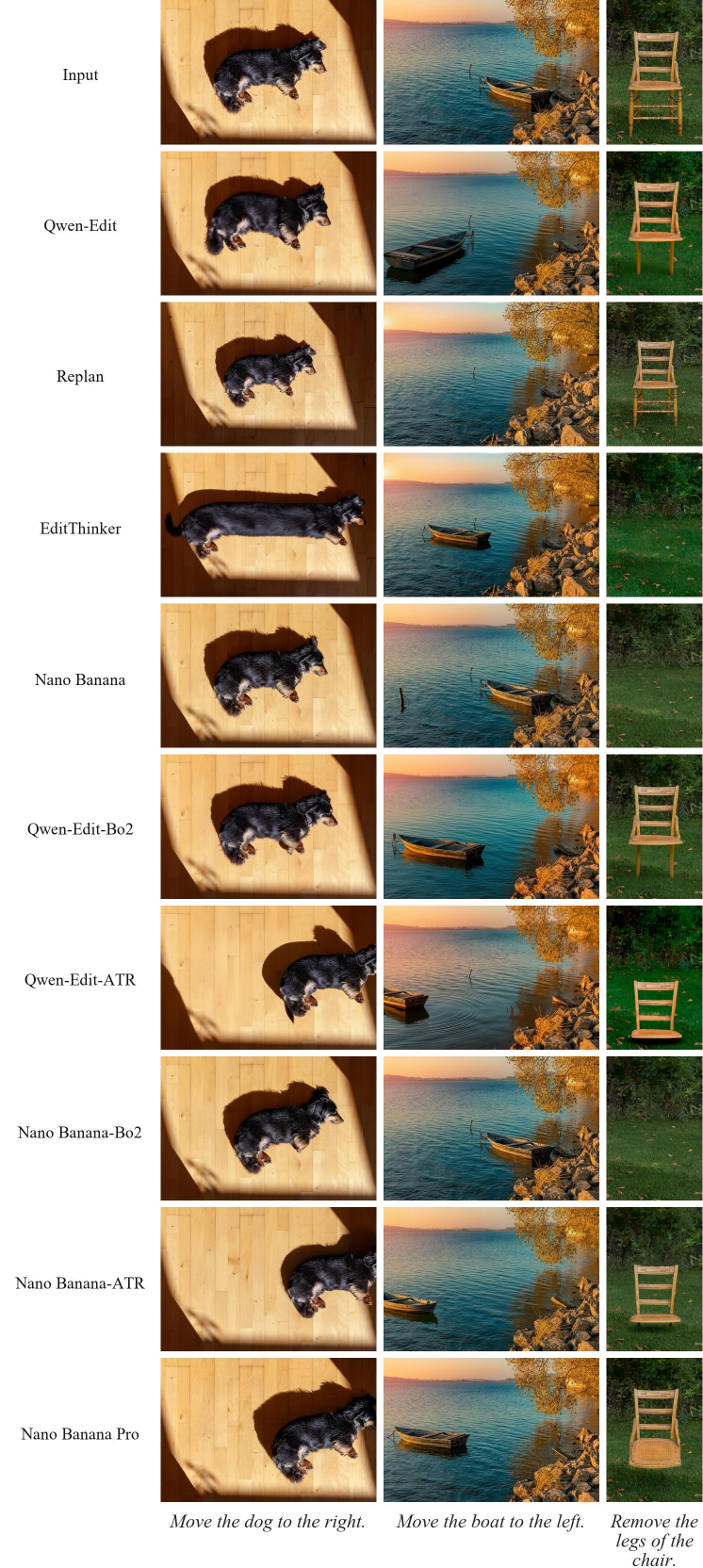}
    \Description{Additional visual comparisons of editing results on the PICA benchmark.}
    \caption{Additional qualitative results on the PICA benchmark.}
    \label{fig:pica_extra}
\end{figure}

Figure~\ref{fig:pica_extra} provides additional qualitative comparisons on the PICA benchmark to demonstrate the generalization capability of our Adaptive Task Reformulation (ATR) framework. These cases highlight the limitations of existing baselines in spatial relocation and physical causal reasoning.

In tasks involving spatial movement, most direct-mapping models struggle with accurate translation, often leaving objects unchanged or generating severe artifacts. For moving the target dog, only Qwen-Edit-ATR, Nano Banana-ATR, and Nano Banana Pro achieve precise relocation. For moving the boat, Qwen-Edit-Bo2 also succeeds alongside these three models, while the rest fail.

For fine-grained editing tasks involving structural dependency and physical causality, most baselines fail to produce meaningful edits or incorrectly erase the entire target object. In contrast, our ATR models and Nano Banana Pro precisely remove the specific parts while accounting for real-world physics, allowing the remaining chair seat to naturally drop to the ground. This demonstrates our framework's advantage in local control and commonsense reasoning.

\subsection{Results on the Replan Benchmark}

\begin{figure}[htbp]
    \centering
    \includegraphics[width=\columnwidth]{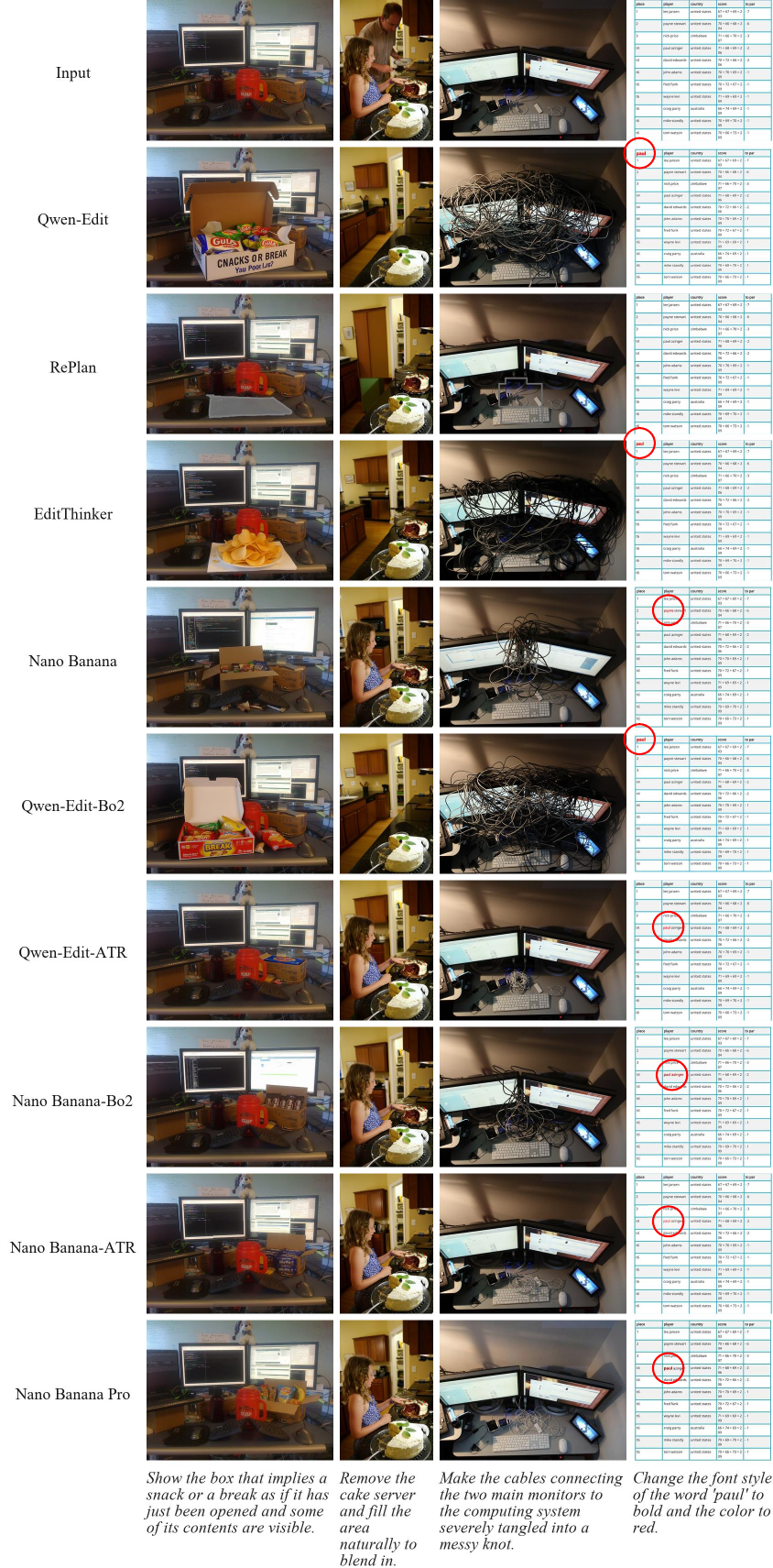}
    \Description{Visual comparisons of editing results on the Replan benchmark.}
    \caption{Additional qualitative results on the Replan benchmark.}
    \label{fig:replan_extra}
\end{figure}

Figure~\ref{fig:replan_extra} further illustrates the performance disparities among models on the Replan benchmark. When instructed to show an opened box, most baselines fail or hallucinate new objects. Conversely, Qwen-Edit-ATR, Nano Banana-ATR, Nano Banana-Bo2, and Nano Banana Pro accurately target and render the opened state. 

In the task of generating tangled cables, baseline models suffer from severe over-editing across the entire desk. Our models successfully restrict editing strictly to the original cable region, generating logical knots. Furthermore, for fine-grained text editing tasks testing pixel-level localization, most models fail due to spatial errors. Qwen-Edit-ATR and Nano Banana-ATR successfully change the target text color to red. Nano Banana Pro further excels by also rendering the required bold effect, demonstrating superior capability in multi-attribute typographic editing.

\subsection{Extended Results on the ImgEdit Benchmark}
\label{sec:appendix_imgedit_wide}

\begin{figure*}[t]
    \centering
    \includegraphics[width=0.80\textwidth]{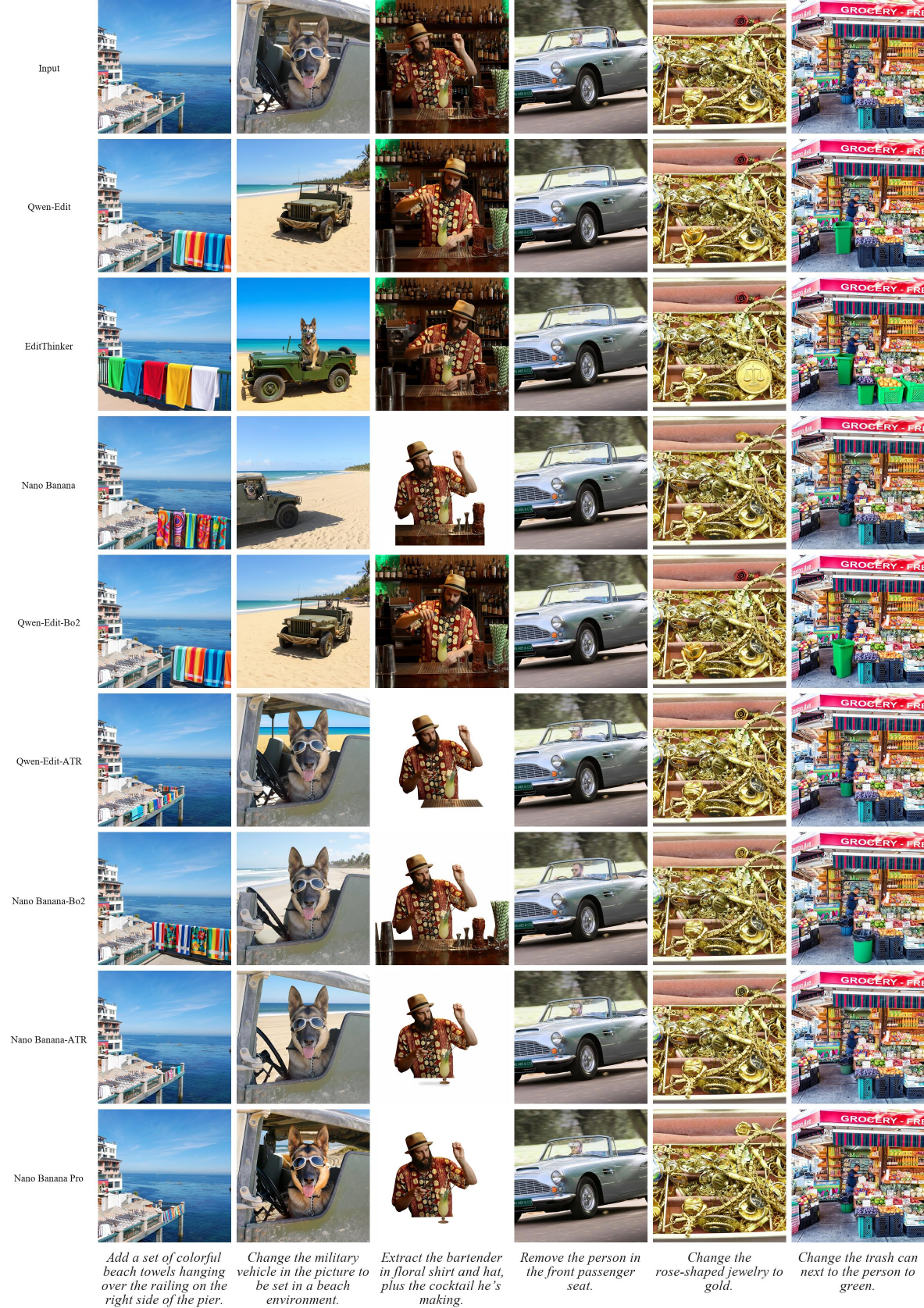}
    \Description{More visual comparisons of editing results on the ImgEdit benchmark.}
    \caption{Additional qualitative results on the ImgEdit benchmark. Our methods (Qwen-Edit-ATR and Nano Banana-ATR) demonstrate consistent improvements across various tasks, achieving performance comparable to Nano Banana Pro model.}
    \label{fig:imgedit_wide_sup}
\end{figure*}

Building upon the qualitative analysis in the main text, Figure~\ref{fig:imgedit_wide_sup} provides a comprehensive set of additional examples from the ImgEdit benchmark. 

Across diverse editing instructions, Qwen-Edit-ATR and Nano Banana-ATR exhibit consistent visual improvements over direct-mapping baselines. In both fine-grained local edits and structural modifications, the ATR framework successfully mitigates common failure modes like over-editing and target ambiguity. These results validate that our approach effectively bridges the performance gap, achieving editing quality comparable to the computationally heavy Nano Banana Pro model.

\section{Detailed Case Studies of Agentic Execution}
\label{sec:appendix_case_studies}

\begin{figure*}[htbp]
    \centering
    \includegraphics[width=0.95\textwidth]{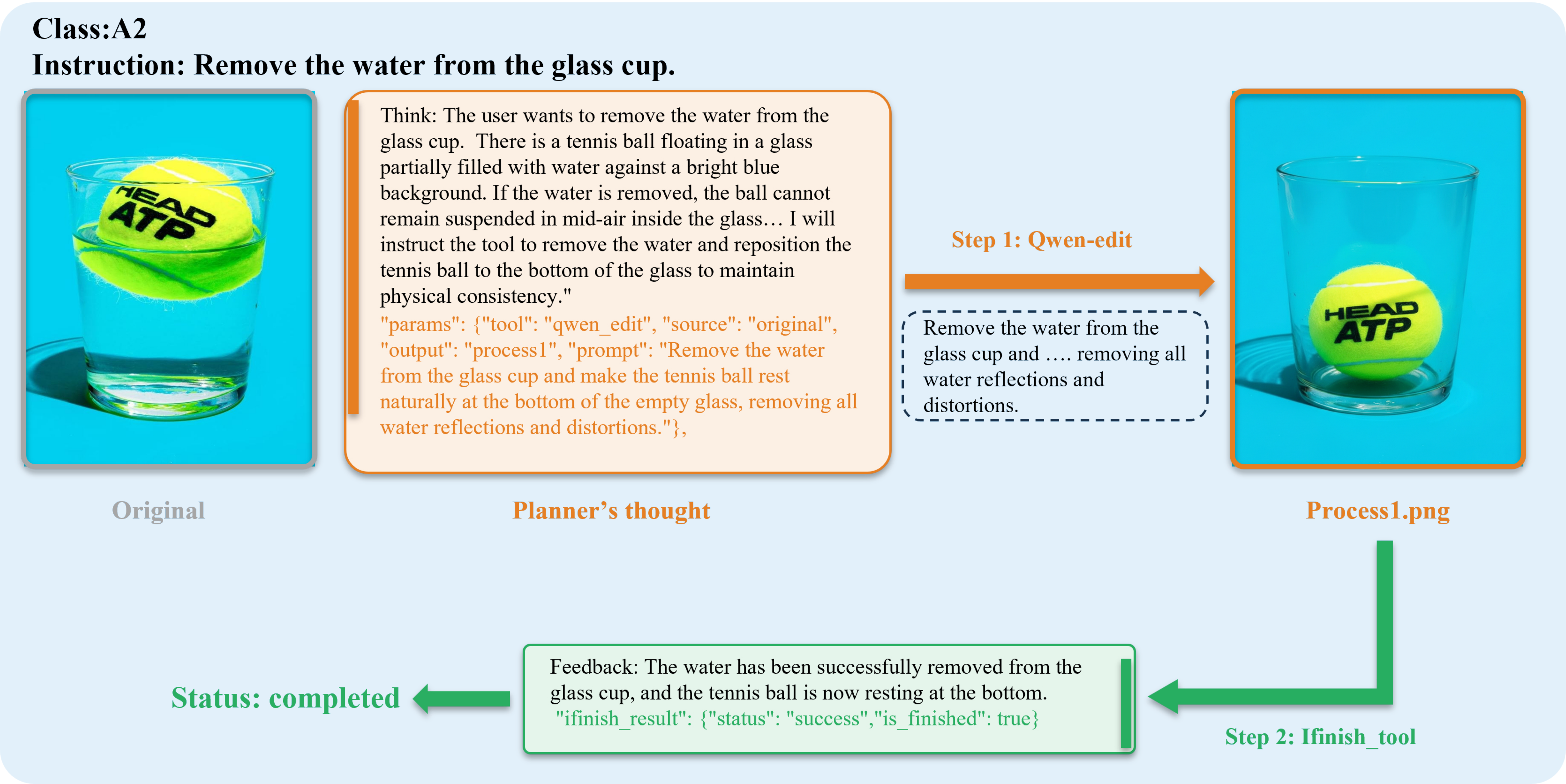}
    \Description{A flowchart detailing the execution steps for the Route A2 strategy.}
    \caption{Detailed execution flow for Route A2 (Instruction Rewriting).}
    \label{fig:case_a2}
\end{figure*}

\begin{figure*}[htbp]
    \centering
    \includegraphics[width=0.95\textwidth]{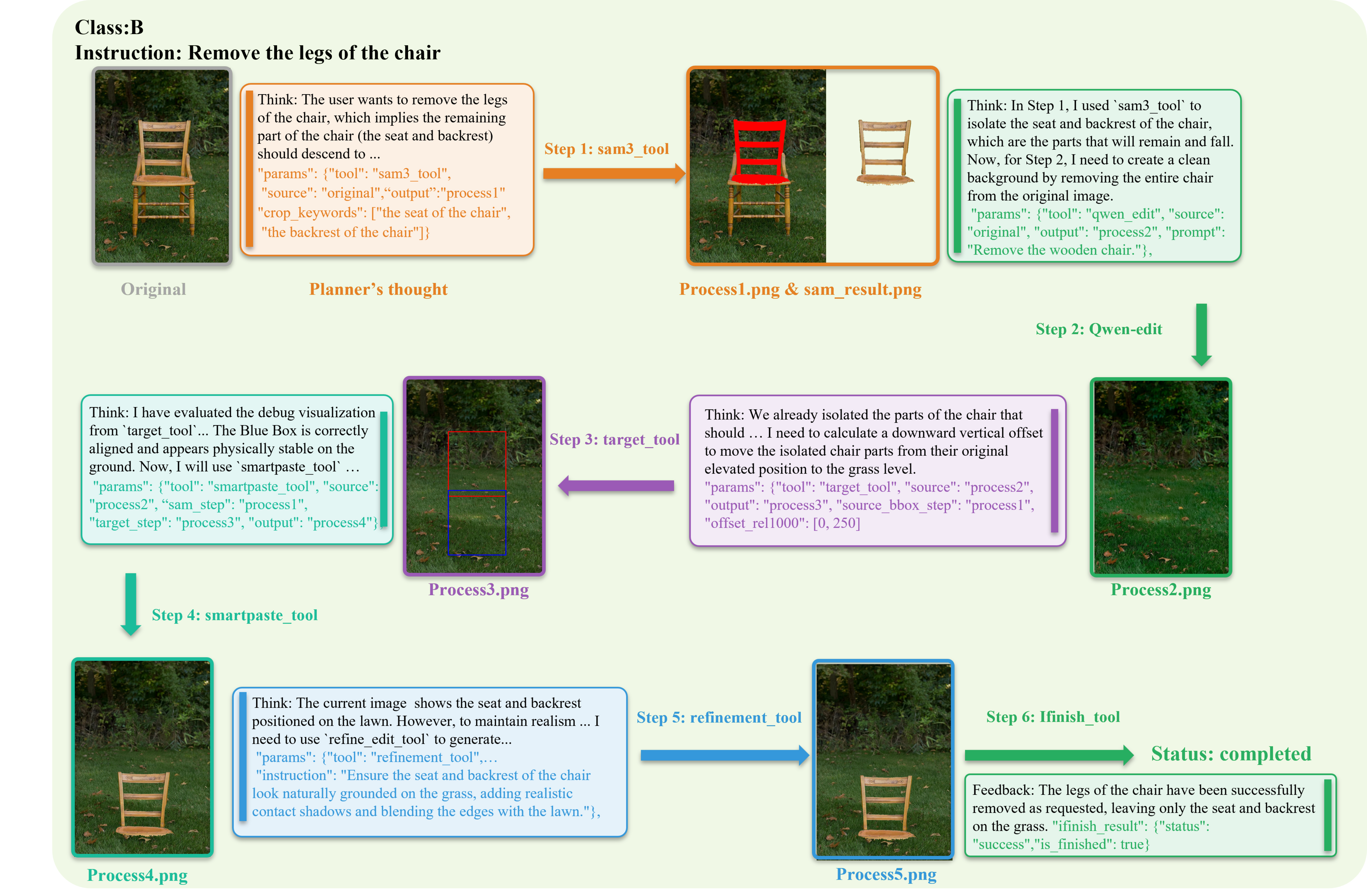}
    \Description{A flowchart detailing the execution steps for the Route B strategy.}
    \caption{Detailed execution flow for Route B (Spatial Decoupling). }
    \label{fig:case_b}
\end{figure*}

\begin{figure*}[htbp]
    \centering
    \includegraphics[width=0.95\textwidth]{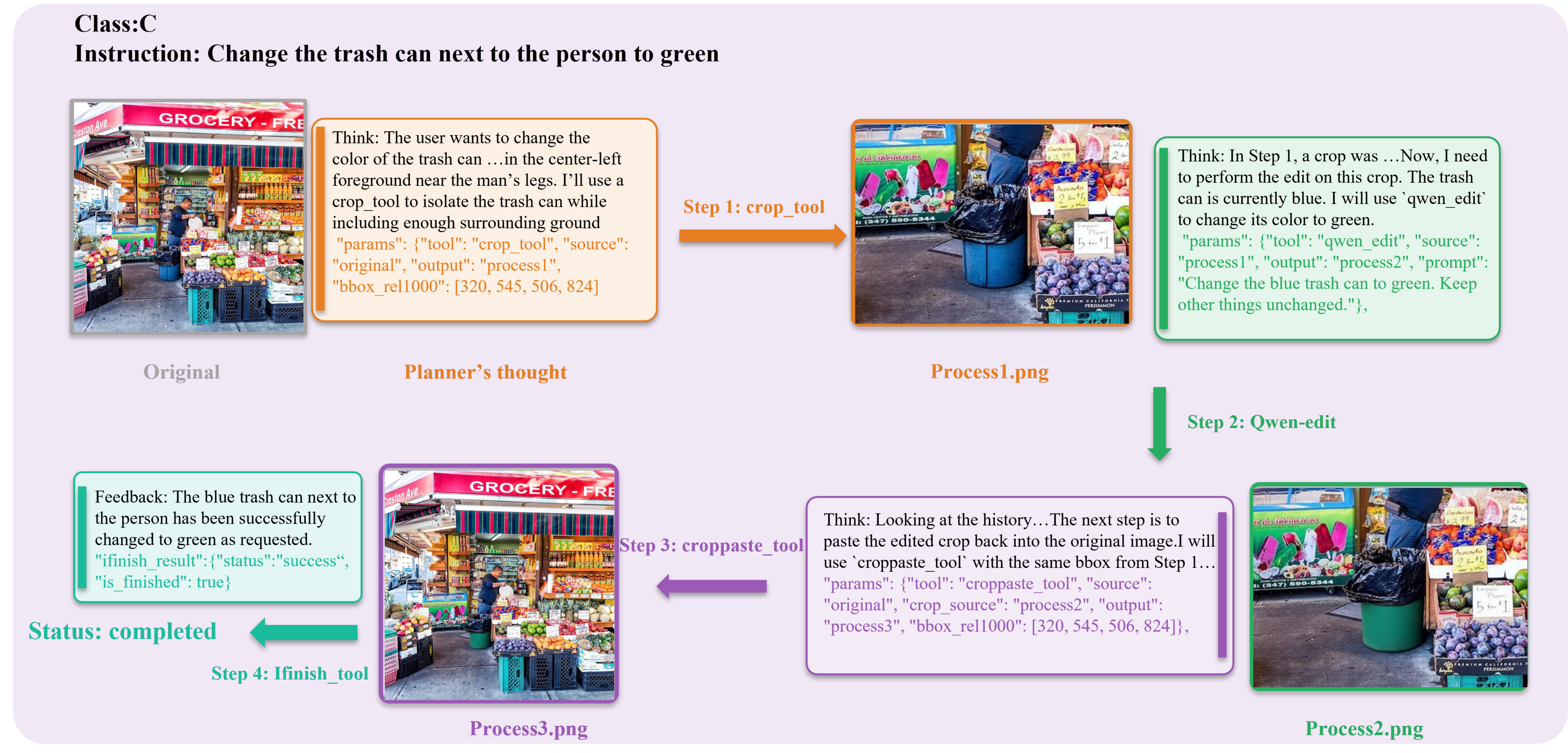}
    \Description{A flowchart detailing the execution steps for the Route C strategy.}
    \caption{Detailed execution flow for Route C (Localized Editing).}
    \label{fig:case_c}
\end{figure*}

To provide a comprehensive and intuitive understanding of the dynamic execution process within our Adaptive Task Reformulation (ATR) framework, this section breaks down the step-by-step reasoning and toolchain invocations of our agent. Specifically, we present three detailed case studies corresponding to our distinct routing strategies:

\begin{itemize}
    \item \textbf{Route A2 (Instruction Rewriting):} Figure~\ref{fig:case_a2} illustrates the execution flow when an ambiguous or reasoning-heavy instruction is intercepted and reformulated into explicit physical actions before being sent to the editor.
    \item \textbf{Route B (Spatial Decoupling):} Figure~\ref{fig:case_b} demonstrates the pipeline for spatial relocation tasks, highlighting how the agent utilizes text-driven segmentation and offset calculation tools to explicitly manage physical relationships.
    \item \textbf{Route C (Localized Editing):} Figure~\ref{fig:case_c} details the execution process for fine-grained local edits, highlighting how the agent isolates the target region to achieve precise modifications while preventing unwanted changes to the global background.
\end{itemize}

\section{Detailed Pilot Study Results on ImgEdit Benchmark}
\label{sec:appendix_pilot_imgedit}

Due to space limitations in the main text, the detailed failure-case analysis and the corresponding recovery performance on the ImgEdit benchmark are presented in this section. Following the identical protocol used for the PICA benchmark discussed in the main text, we sampled the bottom tier of failure cases from Qwen-Image-Edit, classified them via an MLLM, and evaluated the performance of four distinct reasoning strategies.

Table~\ref{tab:bad_case_analysis_pica} presents the distribution of failure categories and the strategy comparison on the ImgEdit benchmark. While the distribution of failure types differs slightly from PICA---with Local entanglement (38.9\%) and Target ambiguity (27.8\%) comprising the majority of difficult cases---the empirical results consistently reinforce the three key findings observed in the main text:

\begin{itemize}
    \item \textbf{Consistent outperformance over direct mapping:} Across every single failure category, direct editing yields the lowest scores. Transforming the task into a more favorable formulation (Rewrite, Spatial, or Local) systematically recovers performance, proving that the bottleneck often lies in the input formulation rather than the base model's absolute capacity.
    \item \textbf{Category-dependent strategy superiority:} The results clearly demonstrate that no single reformulation strategy is a silver bullet. For instance, Localized Editing (Local) significantly excels in resolving \textit{Target ambiguity}, \textit{Local entanglement}, and \textit{Hidden-content reconstruction}, as these tasks heavily rely on high-resolution focus and localized signal-to-noise ratio. Conversely, Spatial Decoupling (Spatial) proves optimal for \textit{Structural dependency} by explicitly managing physical relationships, while Instruction Rewriting (Rewrite) is highly effective for \textit{Transformation complexity}.
    \item \textbf{Robustness of the taxonomy:} The predefined reformulation strategies successfully cover the diverse failure modes presented in the ImgEdit benchmark, with the unclassified ``Others'' category accounting for 0\% of the sampled hard cases.
\end{itemize}

Overall, the stark contrast in strategy effectiveness across different failure categories provides strong empirical validation for our ATR framework: achieving stable and superior image editing requires an intelligent, adaptive system capable of routing each specific instruction to its most suitable reformulation paradigm.

\begin{table}[h]
\centering
\caption{Failure-case analysis across different reasoning strategies on the ImgEdit benchmark. The best result in each row is highlighted in \textbf{bold}.}
\label{tab:bad_case_analysis_pica}
\renewcommand{\arraystretch}{1.12}
\resizebox{\columnwidth}{!}{%
\begin{tabular}{lccccc}
\toprule
\multirow{2}{*}{Failure category} & \multirow{2}{*}{\makecell{Count \\ (\%)}} & \multicolumn{4}{c}{Reasoning strategy} \\
\cmidrule(lr){3-6}
& & Direct & Rewrite & Spatial & Local \\
\midrule
Target ambiguity              & 5 (27.8\%) & 1.20 & 2.80 & 2.60 & \textbf{3.20} \\
Local entanglement            & 7 (38.9\%) & 2.43 & 2.57 & 2.71 & \textbf{3.29} \\
Structural dependency         & 2 (11.1\%) & 2.50 & 3.00 & \textbf{4.00} & 3.00 \\
Hidden-content recons. & 2 (11.1\%) & 3.00 & 4.50 & 3.00 & \textbf{5.00} \\
Scene-wide consistency        & 1 (5.6\%)  & 1.00 & \textbf{3.00} & 2.00 & \textbf{3.00} \\
Transformation complex.     & 1 (5.6\%)  & 1.00 & \textbf{3.00} & 1.00 & 2.00 \\
Others                        & 0 (0.0\%)  & 0.00 & 0.00 & 0.00 & 0.00 \\
\bottomrule
\end{tabular}%
}
\end{table}
\section{Tool Definitions}
\label{sec:appendix_tools}

To facilitate reproducibility, we detail the implementation of our framework, including the definitions, input/output formats, and core mechanisms for all underlying tools utilized during the agentic execution.

Our Adaptive Task Reformulation (ATR) framework relies on a robust library of agent-callable tools. To ensure execution stability and prevent pipeline crashes during automated inference, all geometric tools operate on a standardized relative coordinate system (\texttt{rel1000}), mapping the image dimensions to a $1000 \times 1000$ continuous space. Below, we detail the core tools, categorized by their roles in the pipeline.
\subsection{Core Evaluation and Refinement Tools}

\noindent\textbf{Terminal State Evaluator (\texttt{ifinish\_tool})} \\
This tool acts as an intelligent Quality Assurance node to determine if the editing task should terminate. It employs a Multimodal Large Language Model (MLLM) to compare the original image, the current intermediate image, and the user instruction.
\begin{itemize}
    \item \textbf{Key Mechanisms:} It utilizes a \textit{Focus Constraint} to ignore minor generative artifacts and strictly evaluate whether the core physical action has been achieved. 
    \item \textbf{Output:} A JSON containing the status, boolean \texttt{is\_finished} flag, and reasoning.
\end{itemize}

\noindent\textbf{Artifact Diagnosis \& Prompt Refinement (\texttt{refinement\_tool})} \\
Designed to bridge mechanical transformations and generative smoothing, this tool analyzes the visual discrepancies between the original background and a newly pasted patch.
\begin{itemize}
    \item \textbf{Key Mechanisms:} It forces the MLLM to focus on integration errors, such as truncated structures, missing shadows, or harsh boundaries. It generates a low-level guidance prompt for downstream diffusion models, enforcing a strict hard-truncation constraint to ensure concise generation.
\end{itemize}

\noindent\textbf{Instruction Rewriting Tool (\texttt{fixprompt\_tool})} \\
This tool converts ambiguous, narrative, or reasoning-heavy user queries into explicit physical actions, significantly improving downstream editor success rates.
\begin{itemize}
    \item \textbf{Key Mechanisms:} It acts as a router: if the instruction is already direct, it passes through. If reasoning is required, it applies a series of linguistic transformations to distill complex narratives into precise, unambiguous physical actions. It is designed for high reliability, utilizing deterministic generation and a zero-crash fallback mechanism to ensure continuous execution.
\end{itemize}

\subsection{Route B: Spatial Decoupling Tools}

\noindent\textbf{Text-Driven Segmentation (\texttt{sam3\_tool})} \\
Based on the SAM 3 architecture, this tool extracts high-fidelity masks and bounding boxes from natural language prompts.
\begin{itemize}
    \item \textbf{Key Mechanisms:} It features Adaptive Inference. For single-target queries, it returns the mask with the highest confidence. For multi-target queries, it applies a score threshold ($> 0.25$) and performs a logical OR fusion across multiple masks to capture complex object groupings. It computes both absolute pixel coordinates and safe \texttt{rel1000} relative coordinates.
    \item \textbf{Output:} Bounding boxes, a transparent cutout image, confidence scores, and visualization overlays.
\end{itemize}

\noindent\textbf{Spatial Offset Calculator (\texttt{target\_tool})} \\
This node calculates the precise destination bounding box for an object relocation task. 
\begin{itemize}
    \item \textbf{Key Mechanisms:} It takes the original bounding box and a relative offset array $[dx, dy]$. It explicitly enforces \textit{Size Inheritance} to preserve the object's original proportions and employs strict Boundary Clamping ($\max(0, \min(1000 - w, x))$) to prevent the target box from overflowing the image canvas.
\end{itemize}

\noindent\textbf{Smart Paste (\texttt{smartpaste\_tool})} \\
This tool seamlessly pastes isolated objects into new coordinate regions.
\begin{itemize}
    \item \textbf{Key Mechanisms:} It automatically scans the Alpha channel to trim excess transparent space, ensuring accurate dimensional calculation. It applies aspect-ratio-preserving scaling using LANCZOS resampling to fit the target bounding box, centers the object, and performs alpha compositing.
\end{itemize}

\subsection{Route C: Localized Editing Tools}

\noindent\textbf{Relative Coordinate Cropping (\texttt{crop\_tool})} \\
Extracts localized workspaces based on \texttt{rel1000} coordinates to increase the effective resolution of small targets.
\begin{itemize}
    \item \textbf{Key Mechanisms:} It integrates automatic coordinate normalization to correct spatial boundary errors and employs an Anti-Degeneration safeguard to prevent dimensional collapse, ensuring valid and stable tensor representations for downstream editing models.
\end{itemize}

\noindent\textbf{Adaptive Mixed-Mode Paste (\texttt{croppaste\_tool})} \\
Re-integrates locally edited patches back into the full-resolution image while maintaining contextual consistency.
\begin{itemize}
    \item \textbf{Key Mechanisms:} It utilizes dynamic \textit{Boundary Detection}. If the cropped patch touches the physical edge of the original image, it triggers a \textit{Hard Paste} mode to prevent boundary gradient overflow. If the patch is strictly internal, it employs OpenCV-based \textit{Poisson Blending} for seamless lighting and color transitions. Also, it includes a fallback mechanism to Hard Paste if the Poisson solver encounters a zero-mask exception.
\end{itemize}

\section{Limitation Analysis}
\label{sec:appendix_limitations}

\begin{figure*}[t]
    \centering
    \resizebox{0.95\textwidth}{!}{%
        \includegraphics{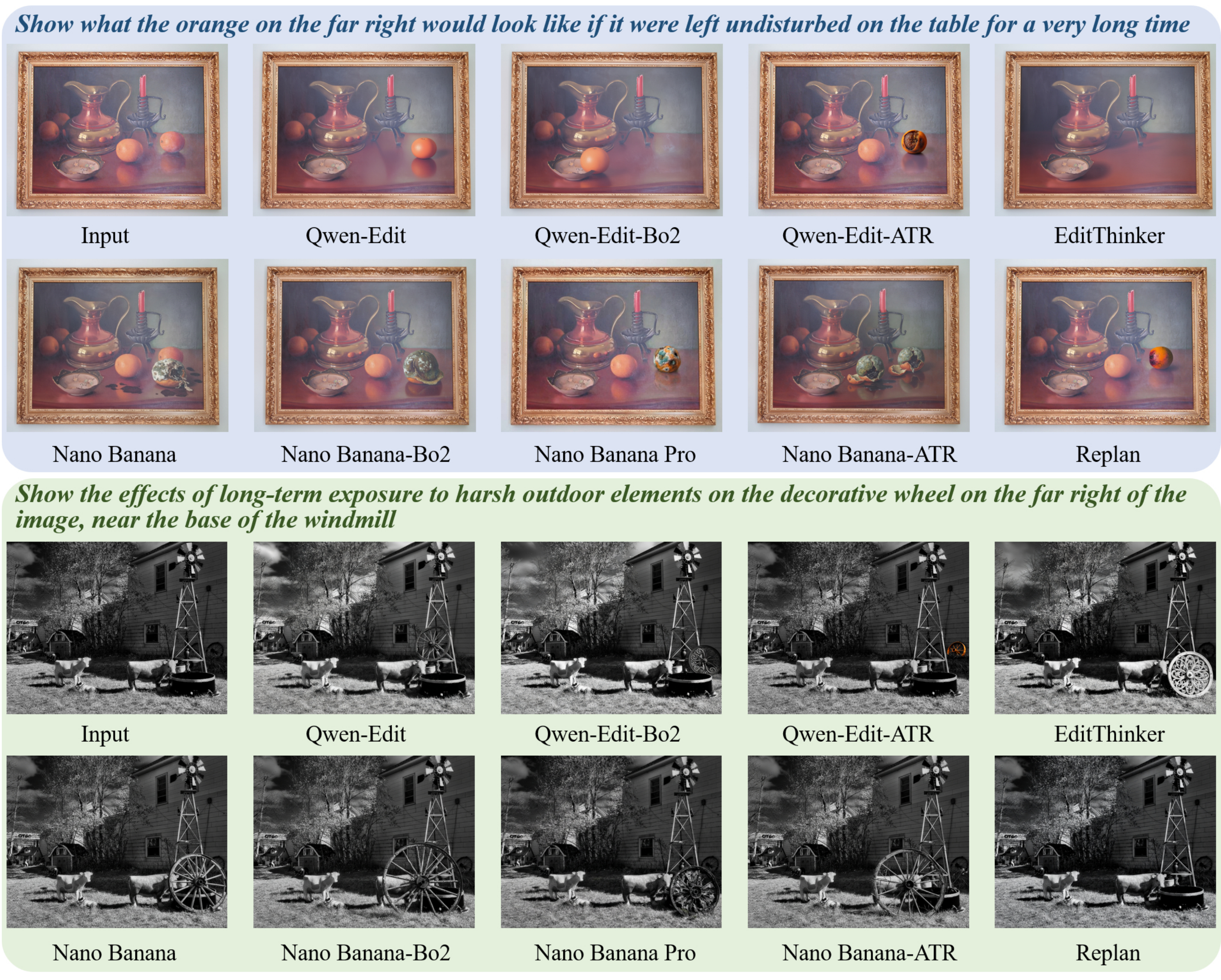} 
    }
    \Description{Images illustrating limitations of the proposed framework.}
    \caption{Limitations of our framework.}
    \label{fig:limitations}
\end{figure*}
In image editing tasks, striking a balance between precise local modifications and global attribute consistency remains a highly challenging sub-task. As shown in Figure~\ref{fig:limitations}, both current state-of-the-art (SOTA) baselines and our Adaptive Task Reformulation (ATR) framework exhibit limitations in such scenarios. The top row illustrates a conflict with global artistic style. When modifying an object within a classical oil painting, baseline models generally fail---either leaving the target unchanged, removing it entirely, or disrupting the spatial layout. Our Qwen-Edit-ATR successfully decays the far-right orange without altering surrounding objects; however, constrained by its localized workspace, the newly generated texture exhibits a noticeable mismatch with the global classical atmosphere. 

Another case of this local-global conflict involves global color constraints, as shown in the bottom row. Here, the original image is strictly grayscale, but the requested effect inherently carries strong color characteristics. Most baselines suffer severe errors, either replacing the wheel entirely or hallucinating duplicates. In contrast, Qwen-Edit-ATR accurately applies a realistic weathering effect strictly to the target wheel while preserving its structure. Yet, this "perfect local edit" exposes a drastic visual dissonance: isolated from the global black-and-white context, the model generates a colored rusty wheel. Recomposing this localized, colored output back into the grayscale scene creates a glaring anomaly, indicating that incorporating global style and attribute awareness into local agentic execution is a crucial direction for future improvement.
\section{Refinement of the PICA Evaluation Protocol}
\label{sec:appendix_pica_refinement}

The standard PICA benchmark evaluates editing performance through a localized Vision Question Answering (VQA) protocol. Specifically, it crops a pre-defined target region from the edited image and poses a specific question to a Vision Large Language Model. If the VLM's answer matches the human-annotated Ground Truth (GT), the edit is deemed successful. 

However, during our empirical analysis, we observed that a subset of these QA pairs exhibits a strong, unreasonable bias. In many open-ended editing scenarios, the benchmark imposes overly rigid or physically illogical deterministic constraints. Consequently, high-quality, physically plausible edits are unjustly penalized simply because they do not conform to these flawed ground truths.

\begin{figure*}[t]
    \centering
    \includegraphics[width=0.95\textwidth]{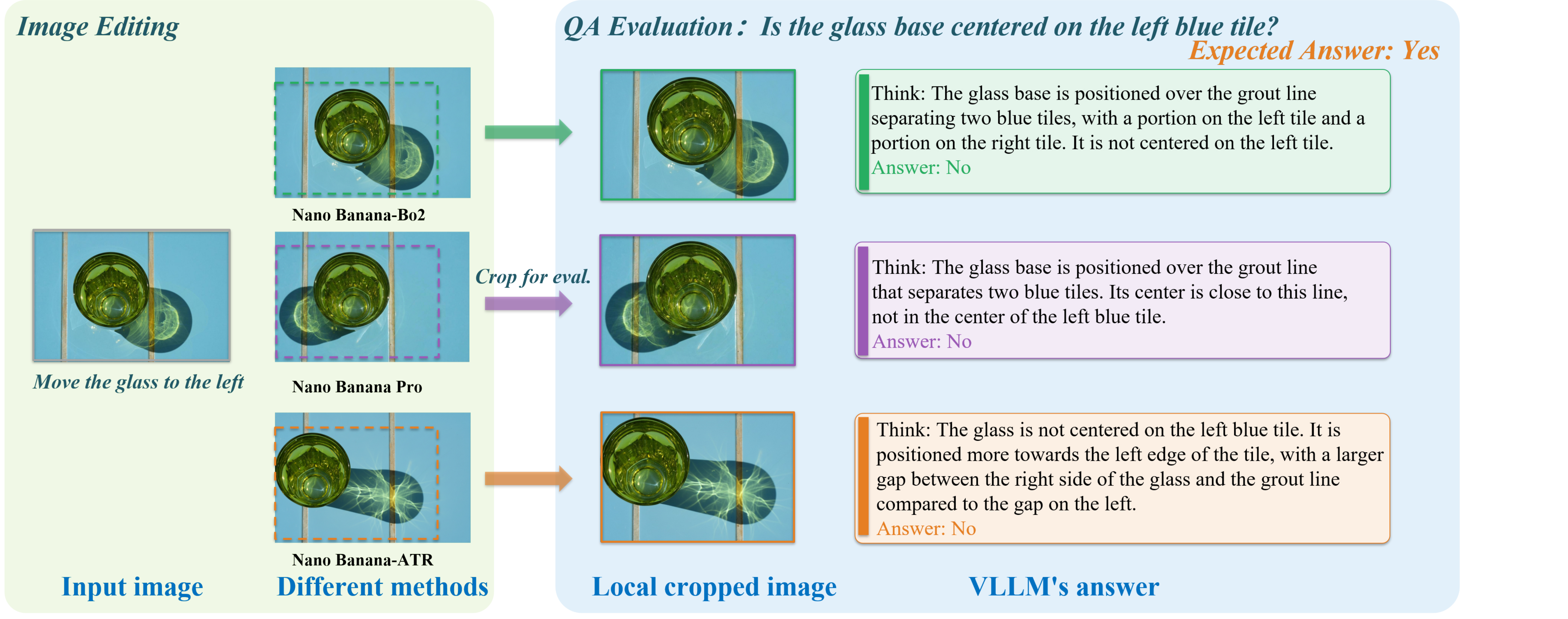}
    \Description{An example image showing a flawed QA evaluation case regarding moving a glass.}
    \caption{Ill-posed QA example 1: Moving the glass. A logically correct edit is penalized due to an unreasonable spatial constraint.}
    \label{fig:pica_flaw_glass}
\end{figure*}

To illustrate this, consider the case of moving an object, as shown in Figure~\ref{fig:pica_flaw_glass}. The instruction is \textit{``Move the glass to the left''}. The corresponding evaluation question asks, \textit{``Is the glass base centered on the left blue tile?''} with an Expected Answer of \textit{``Yes''}. However, given the original layout, strictly centering the glass on the left tile would force a significant portion of the glass outside the image boundaries, which violates common sense. Models like Nano Banana Pro and our Nano Banana-ATR correctly translate the glass to the left while keeping it entirely within the frame. The VLM accurately observes the scene and answers \textit{``No''}, resulting in an unfair false negative.

\begin{figure*}[t]
    \centering
    \includegraphics[width=0.95\textwidth]{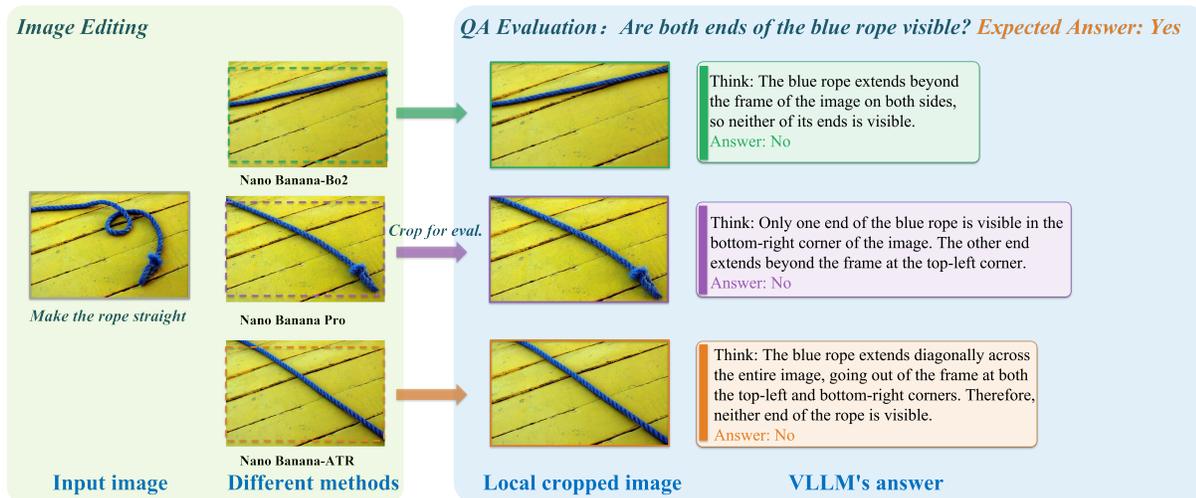}
    \Description{An example image showing a flawed QA evaluation case regarding straightening a rope.}
    \caption{Ill-posed QA example 2: Straightening the rope. A successful structural edit fails the evaluation because the natural length extends beyond the frame.}
    \label{fig:pica_flaw_rope}
\end{figure*}

A similar structural bias is observed in Figure~\ref{fig:pica_flaw_rope}. The instruction is \textit{``Make the rope straight''}, with the evaluation question asking, \textit{``Are both ends of the blue rope visible?''} (Expected Answer: \textit{``Yes''}). Both our ATR variant and the Pro model successfully execute the instruction, rendering a perfectly straight rope. However, due to the natural length of the rope and the fixed camera perspective, straightening it inevitably causes the ends to extend beyond the image boundaries. The VLM correctly answers \textit{``No''}, once again penalizing models that successfully completed the core task but failed an arbitrary secondary constraint.

To ensure a rigorous and fair assessment of true editing capabilities, we employed a Multimodal Large Language Model (MLLM) to systematically screen and filter out these inherently flawed or unreasonably biased QA pairs. Following this data refinement phase, the total number of valid QA pairs was reduced from 4,112 to 3,879. This refined evaluation subset prevents models from being penalized for maintaining real-world logic, providing a much more objective comparison for physical realism.
\end{document}